\documentclass[hidelinks,11pt]{article}

\usepackage[stable]{footmisc}

\usepackage{adjustbox}
\usepackage{amsmath}
\usepackage{amssymb}
\usepackage{caption}
\usepackage{comment}
\usepackage{dirtytalk}
\usepackage{float}
\usepackage[T1]{fontenc}
\usepackage{mathptmx}
\usepackage{graphicx}
\usepackage[utf8]{inputenc}
\usepackage{csquotes}
\usepackage{hyperref}
\usepackage{mathtools}
\usepackage{multicol}
\usepackage{multirow}
\usepackage[round]{natbib}
\usepackage[all]{nowidow}
\usepackage{soul}
\usepackage{subcaption}
\usepackage{tipa}
\usepackage[normalem]{ulem}
\usepackage{wasysym}
\usepackage[svgnames]{xcolor}
\usepackage{subcaption}
\usepackage[paperheight=27.94cm,paperwidth=21.59cm,left=3.81cm,right=3.81cm,top=3.81cm,bottom=3.81cm]{geometry}
\usepackage{upgreek}
\usepackage{xurl}
\usepackage{CJKutf8}
\usepackage{cjhebrew}
\usepackage{stmaryrd}
\usepackage{array}
\usepackage{booktabs}

\newcommand{\zh}[1]{\begin{CJK}{UTF8}{gbsn}#1\end{CJK}}

\title{Translation in the Wild\footnote{Final version published in \textit{Information}: \href{https://doi.org/10.3390/info16121077}{doi.org/10.3390/info16121077}.}}
\date{}
\author{Yuri Balashov\\Department of Philosophy\\Institute for Artificial Intelligence \\University of Georgia\\Athens, GA 30602, USA \\ \href{mailto:yuri@uga.edu}{yuri@uga.edu} \\ ORCID ID: 0000-0001-7369-2122}

\begin{document}

\maketitle

\begin{abstract}
\noindent Large Language Models (LLMs) excel in translation among other things, demonstrating competitive performance for many language pairs in zero- and few-shot settings. But unlike dedicated neural machine translation models, LLMs are not trained on any translation-related objective. What explains their remarkable translation abilities? Are these abilities grounded in “incidental bilingualism” \citep{briakou_searching_2023} in training data? Does instruction tuning contribute to it? Are LLMs capable of aligning and leveraging semantically identical or similar monolingual contents from different corners of the internet that are unlikely to fit in a single context window? I offer some reflections on this topic, informed by recent studies and growing user experience. My working hypothesis is that LLMs’ translation abilities originate in two different types of pre-training data that may be internalized by the models in different ways, \emph{Local} and \emph{Global}. “Local learning” makes use of bilingual signals present within a single training context window (e.g. an English sentence followed soon by its Chinese translation in the training data). “Global
learning,” in contrast, capitalizes on mining semantically related monolingual contents that are spread out over the LLMs' pre-training data. The key to explaining the origins of LLMs' translation capabilities is a continuous iteration between Local and Global learning, which is a natural and helpful consequence of batch training. I discuss the prospects for testing the “duality hypothesis” empirically and its implications for reconceptualizing translation, human and machine, in the age of deep learning.

\par
\vspace{3mm}

\end{abstract}

\newpage

\tableofcontents

\section{Introduction}\label{intro}

Large Language Models’ (LLMs) relation to translation is curious, both historically and conceptually. Historically, all the landmark achievements leading up to generative AI were made in the context of machine translation (MT), in the space of three years. This includes the original sequence-to-sequence model \citep{sutskever_sequence_2014}, the classical attention mechanism \citep{bahdanau_neural_2014, luong_effective_2015} and the transformer \citep{vaswani_attention_2017}, as well as the attendant tokenization algorithms \citep{sennrich_neural_2016, kudo_sentencepiece_2018}. Introduced initially to improve the state of the art in MT by 2 or 3 BLEU points,\footnote{BiLingual Evaluation Understudy \citep{papineni_bleu_2001}---a popular metric for automatic evaluation of MT quality based on an averaged $n$-gram (usually up to $n = 4$) overlap between an MT output and a reference translation.} these achievements were immediately adopted in virtually every other area of deep learning, from text to image to video generation and more.

From a more theoretical perspective, it is not so clear why LLMs translate as well as they do, showing competitive results on standard benchmarks for many language pairs in zero- and few-shot settings.\footnote{For a recent comprehensive overview, see \citealp{ataman_machine_2025}.} Unlike dedicated neural MT models, LLMs are not trained on any translation-related objective. Indeed, they are pre-trained almost exclusively on English---basically the entire Anglocentric Internet---with only a small percentage of non-English content \citep{blevins_language_2022, lin_few-shot_2022}. Translation is, therefore, an \emph{emergent} ability of LLMs alongside reasoning, text summarization and generation, and so forth \citep{wei_emergent_2022}. What explains its origin?

The sheer size of LLMs (trillions of parameters) and of their pre-training data (hundreds of billions of tokens), as well as the presence of some multilingual content in it (on the order of 3--10\% in most cases) are important factors. In comparison, a strong transformer-based neural machine translation (NMT) model requires anywhere from 100M to 10B parameters and a comparable number of aligned sentence pairs in two languages to train to perfection. A three-orders-of-magnitude difference in the model and data size, along with an admixture of multilingual content, suggests that LLMs could probably internalize everything anyone has ever translated, as long as it ended up on the internet in one form or another \citep{morris_how_2025}.

But this is just a starting point. While multilingual pre-training data is key to explaining the translation abilities of LLMs, that data is very heterogeneous in form and content. LLMs may have been exposed to generally available parallel corpora (such as the OPUS collection, WMT benchmarks, United Nations proceedings, and other open repositories of multilingual texts often used to train dedicated NMT engines) during pre-training. They may also have seen foreign language teaching texts, bilingual abstracts of scientific papers, and other trace amounts of “incidental bilingualism”---“needles in a haystack,” as \cite{briakou_searching_2023} put it. And in the last two years they may have learned more from translation queries made to the free versions of the largest models.

But this alone can hardly explain the remarkable translation performance of LLMs, for at least two reasons: (i) the corresponding sentences in two languages (e.g. English and German) in online corpora, textbooks, and so forth, are usually separated by intervening text. They do not follow one another, making them suboptimal for an autoregressive language modeling objective. (ii) Parallel sentences are present only in a very small proportion of the total multilingual pre-training data used by LLMs.

Most of the multilingual content in LLM training data comes in the form of monolingual documents found in different parts of the internet. One fully expects these documents (e.g. news articles or Wikipedia pages in different languages) to include semantically identical or very similar sentences across languages. But they do not arrive in neatly aligned pairs and are unlikely to co-occur within a single context window used in pre-training (4–100K tokens at the time of writing). And even when such parallel or near-parallel sentences do co-occur in a common context window, they are typically separated and surrounded by a great deal of “noise.”

It is hard to resist the idea that contemporary LLMs must somehow be able to locate and leverage semantically related monolingual content from the noisy multilingual soup of their training data, and that this ability contributes to their translation performance and other multilingual abilities such as responding to questions or instructions in different languages (see, in particular, \citealp{chua_crosslingual_2023, gao_understanding_2025}). If that is indeed the case, then the translation abilities of LLMs may have \emph{two distinct sources} and may learn from these sources in two  somewhat different ways. For the sake of discussion, I will refer to these as \emph{Local} and \emph{Global}. “Local learning” makes use of bilingual signals present within a single context window (e.g. an English sentence followed soon by its French translation in the training data). “Global learning,” in contrast, capitalizes on making the best use of semantically related monolingual contents that are spread out over the LLM’s pre-training data.

‘Local’ and ‘Global’ are rough-and-ready labels, and I offer them primarily as a heuristic framework for exploration of potentially fruitful connections. Due to differences in their architecture, training data, and training objectives, dedicated NMT systems and LLMs perform translation tasks in significantly different ways (see Section~\ref{NMT-vs-LLMs}). If, in addition, LLMs \emph{learn} to translate in two different ways, Local and Global, then neural-network-based translation in a broader sense (encompassing both traditional NMT models and LLMs prompted for translation tasks) cannot be conceptualized as a unified process. Perhaps this is not surprising given the checkered history of MT and the variety of approaches to translation developed along the way (Section \ref{brief-history-trans-tech}), each stage introducing a new understanding of what it means, and takes, to translate. What \emph{is} different nowadays is the stunning translation quality brought by neural networks and deep learning, which makes the need for understanding their inner workings more pressing.

Notably, \emph{human translation} is far from being a univocal concept either. This is obvious from a cursory glance at the landscape of modern translation studies \citep{snell-hornby_turns_2006, venuti_translation_2013, munday_introducing_2022} highlighting the complexity and subjectivity of human translation and its dependence on various linguistic, cultural, cognitive, environmental, and ergonomic factors. Recent progress in computer-assisted human translation technologies has added even more complexity to the picture \citep{rothwell_translation_2023, moorkens_automating_2025}. At the end of the day, we humans may translate in more than one way, especially when we are equipped with CAT (computer-assisted translation) tools which extend our cognitive abilities in non-trivial manners \citep{balashov_translators_2020}, and artificial neural networks may do it differently from us, and even from each other. The overall lesson from the recent developments in translation technologies, both human and machine, is that there may be no such thing as Translation with a capital ‘T’.

The primary goal of this paper, however, is to reflect on the origins of the remarkable translation abilities of LLMs in light of recent research and growing user experience. The plan is as follows. Section \ref{brief-history-trans-tech} provides a brief overview of the history of translation technologies with emphasis on the latest developments. Section \ref{NMT-vs-LLMs} compares standard state-of-the-art NMT models with LLMs. (Readers familiar with the material of Sections \ref{brief-history-trans-tech} and/or \ref{NMT-vs-LLMs} may skip ahead.) Section \ref{recent-uses-of-LLMs} reviews recent applications of LLMs to translation and translation-related tasks. In Section \ref{how-llms-translate}, I discuss important theoretical work exploring \emph{how} LLMs perform these tasks. In Section \ref{what-explains}, I turn to the question of \emph{why} LLMs are so good at them and develop a “dualistic” proposal about the origin of LLMs’ translation abilities in two types of pre-training data and two learning processes. Section \ref{prospects-empirical} considers prospects for operationalizing and testing this proposed duality empirically. Section \ref{implications-concept-translation} draws broader implications for our rapidly evolving concept of translation in the deep learning era. Section \ref{conclusion} offers concluding remarks.

\section{Translation Technologies: Brief History and Current State} \label{brief-history-trans-tech}

\begin{figure}[h]
    \centering
    \includegraphics[width=1.0\textwidth]{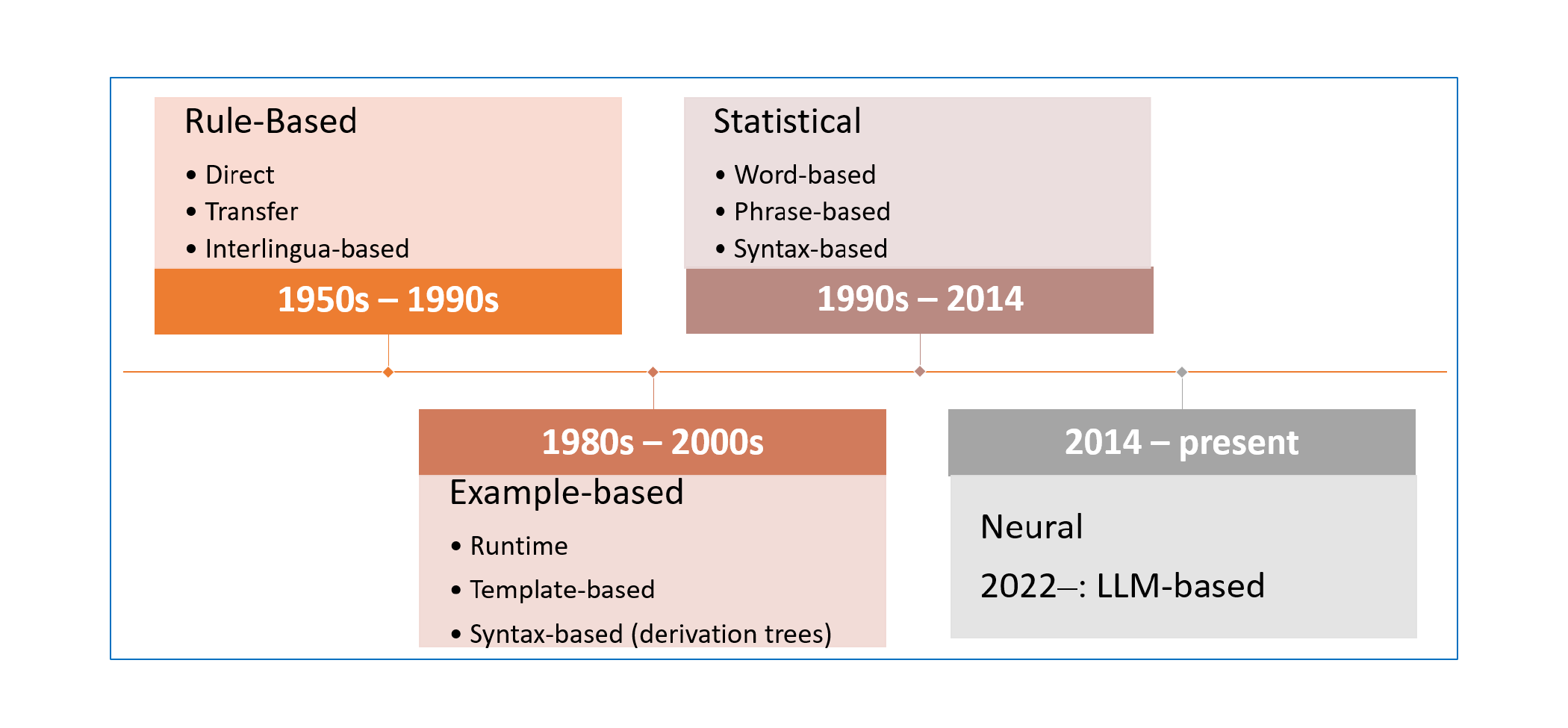}
    \caption{Brief history of translation technologies}
    \label{fig:figure1}
\end{figure}

The development of translation technologies has a rich history, with multiple paradigm shifts over the past decades (Figure~\ref{fig:figure1}). Each stage in this development  introduced new techniques as well as new conceptualizations of what translation entails. For useful historical accounts, the reader is referred to \citet{hutchins_machine_1986, jurafsky_speech_2009, sin-wai_routledge_2015, koehn_neural_2020, mercan_evolution_2024, ataman_machine_2025}.

Today’s translation technology landscape is one of convergence and integration \citep{rothwell_translation_2023, moorkens_automating_2025}. The boundary between human and machine translation is rapidly blurring. Professional translators use computer-assisted translation (CAT) tools that are often integrated with a custom NMT system or a generic service such as DeepL or Google. The translator’s role is often to post-edit MT output and curate translation memories, a process that loops human expertise back into improving machine suggestions. This symbiosis has led to significant productivity gains and reconceptualization of translation as a joint human-AI effort \citep{balashov_translators_2020}, where creativity and critical judgment come from humans, and speed and consistency from machines.

Most recently, the emergence of LLMs has begun to influence translation workflows. Although NMT systems are still the go-to resource for high-volume, high-speed translation needs \citep {makarenko_list_2025, big_language_solutions_why_2025}, LLMs have demonstrated impressive translation capabilities even without being specialized for the task. There are early signs of LLMs being integrated into translation pipelines; for instance, using an LLM to refine or evaluate MT outputs, or to handle difficult cases that NMT systems mistranslate. Conversely, practitioners have explored using NMT engines to augment LLMs, for example, by translating non-English user queries into English before feeding them to an English-centric LLM, and then translating the output back to the original language using a domain-specific MT system.

The rapidly growing literature on using LLMs in translation tasks reveals that, with strategic prompting, LLMs can perform progressively sophisticated operations including, but not limited to: evaluating the quality of translation output, including their own \citep{kocmi_large_2023, fernandes_devil_2023, lu_error_2024}; spotting and categorizing translation errors and suggesting corrections \citep{berger-etal-2024-prompting, feng_tear_2024}; automatic post-editing of raw MT output \citep{raunak_leveraging_2023, ki_guiding_2024, alves_tower_2024}; adapting translation output to the specific terminology \citep{ghazvininejad_dictionary-based_2023, rios_instruction-tuned_2024}, to a given domain (e.g. pharmaceuticals, oil and gas, IT) \citep{sia_-context_2023, zheng_fine-tuning_2024}, and to existing translation memories and other project-, client- or domain-specific instructions and reference materials, often outperforming in these respects more traditional approaches earlier implemented in NMT systems \citep{moslem_adaptive_2023, moslem_language_2024, vieira_how_2024}; generating glossaries of special terms from pairs of source and target documents \citep{ding_enhancing_2025}; improving the quality of translation in low-resource directions (e.g. Swahili-Japanese) by following a “chain-of-thought” \citep{wei_chain--thought_2022} prompt which explicitly requires them to pivot (“Translate this sentence from Swahili to English first; then translate the English output to Japanese,” see Section \ref{pivoting-intermadiate}); following, with benefit, a human translation workflow \citep{chen-etal-2024-iterative, he_exploring_2024}, often by engaging LLMs in a prolonged interaction involving pre-translation research, drafting, refining, and proofreading \citep{briakou_translating_2024}. The opportunities in this area are virtually unlimited.

Let’s turn to a closer comparison of dedicated NMT models and LLMs as agents of translation, to clarify how they differ and what those differences imply.

\section{NMT vs LLMs: A Comparison} \label{NMT-vs-LLMs}

Dedicated Neural Machine Translation (NMT) models and Large Language Models (LLMs) share a common foundation: both utilize transformer-based neural architectures optimized for autoregressive text generation. Yet they differ profoundly in architecture details, training data, learning objectives, and modes of usage. These differences illuminate the distinctions critical for understanding LLMs’ translation capabilities, especially in contrast to standard NMT systems.

\subsection{Model Architecture and Size}

\begin{figure}[ht]
    \centering
    \includegraphics[width=1.0\textwidth]{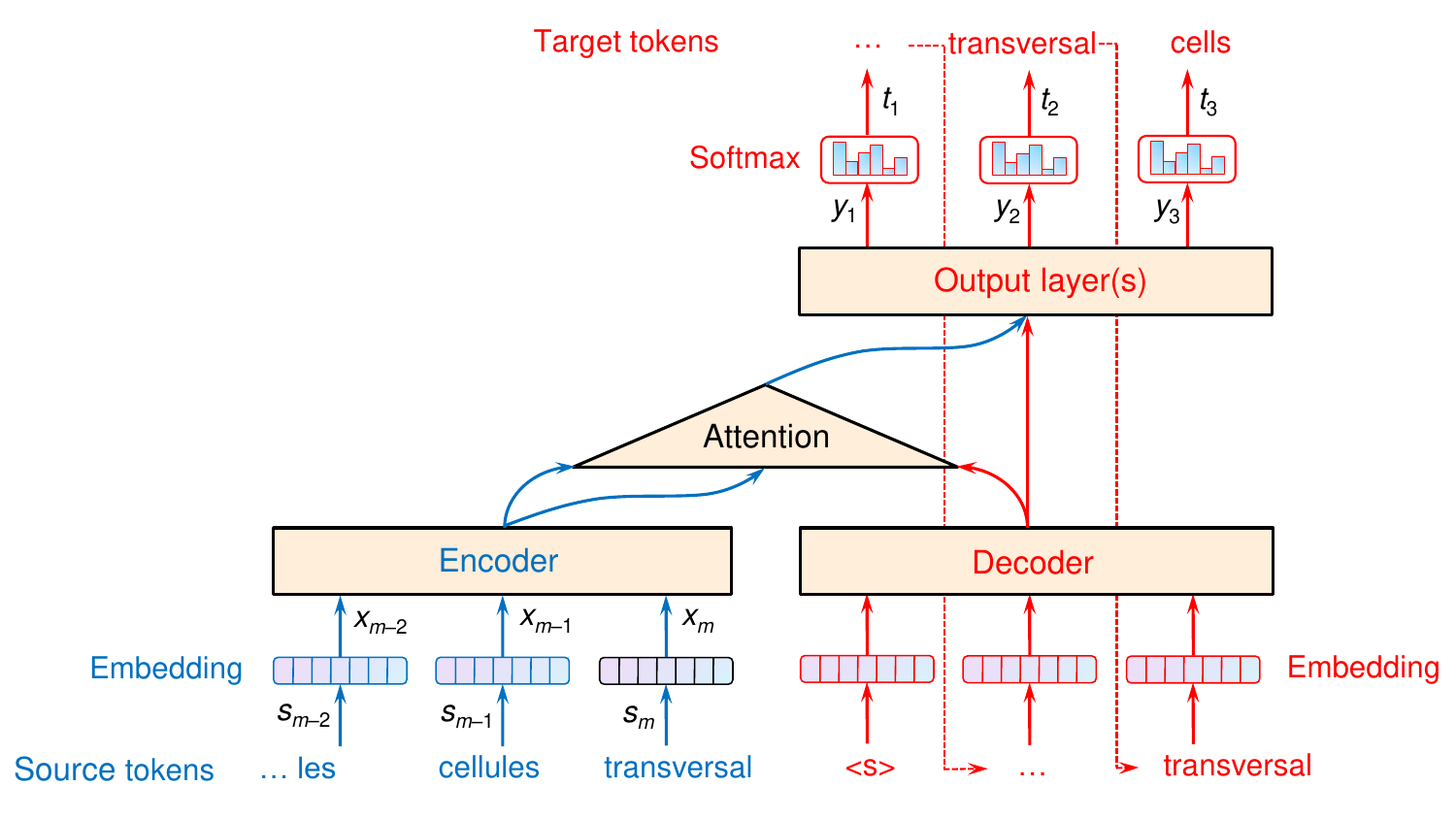}
    \caption{NMT architecture}
    \label{fig:figure2}
\end{figure}

\begin{figure}[ht]
    \centering
    \includegraphics[width=0.75\textwidth]{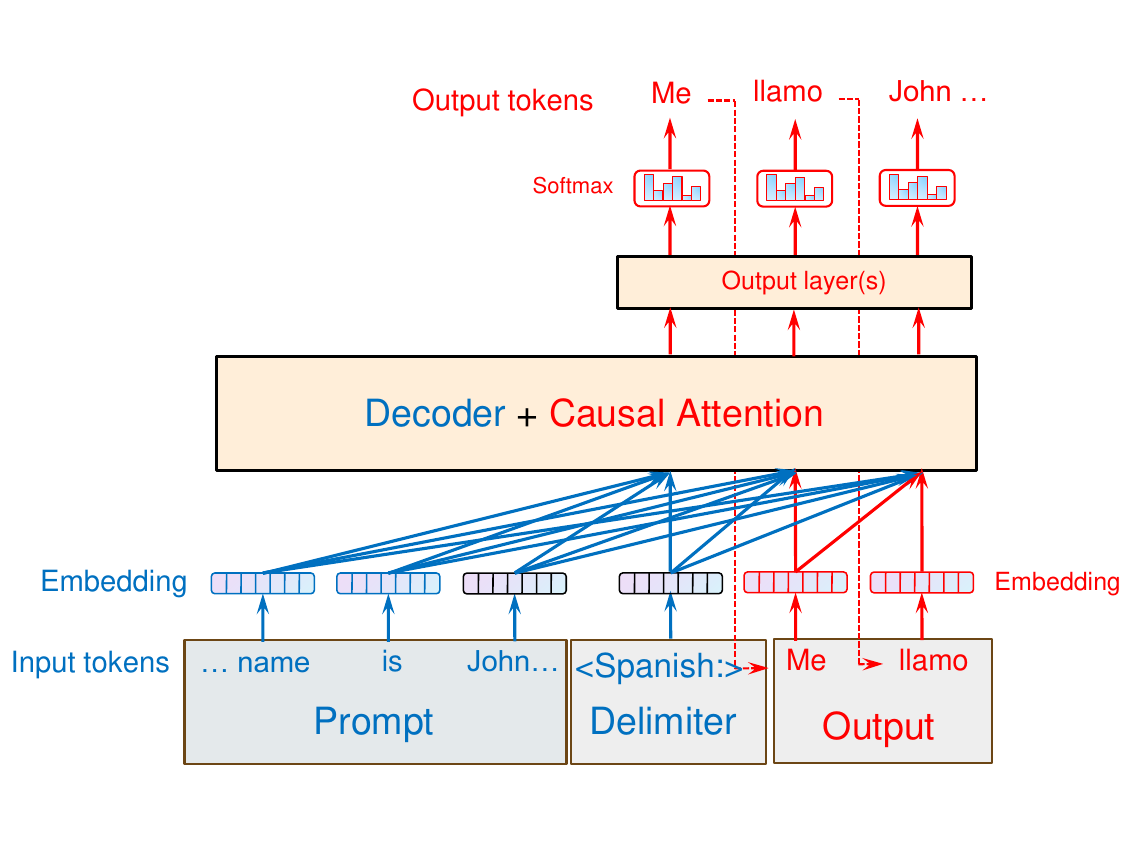}
    \caption{LLM architecture}
    \label{fig:figure3}
\end{figure}

Typical state-of-the-art NMT models have 100M to 10B parameters and 6–12 transformer layers on the encoder and decoder sides. In contrast, cutting-edge LLMs are rumored to have \emph{trillions} of parameters (two to three orders of magnitude larger). NMT models are usually \emph{encoder-decoder} transformers: an encoder processes the source sentence, and a decoder generates the target sentence, attending to the encoder’s representation (Figure~\ref{fig:figure2}). Most LLMs, on the other hand, are \emph{decoder-only} transformers, especially those designed for text generation. A decoder-only model produces output based on a single stream of input, which can include instructions or source text (Figure~\ref{fig:figure3}). Despite this difference, an LLM can effectively act like an encoder-decoder when the prompt includes the source text: the instruction implicitly tells it to “encode” the source meaning into its latent state before producing the output; for example: \texttt{\small{Translate the following sentence to Spanish: “My name is John.”}}

\subsection{Training Data}

This is perhaps the starkest difference. An NMT model is trained on a relatively narrow and specific dataset: a large collection of \emph{parallel corpora} (source-target sentence pairs). For example, a high-quality English-German NMT model might be trained on tens of millions of aligned sentence pairs from sources like EU Parliament proceedings, translated news, subtitles, and so forth. The NMT model thus directly learns to map between two languages. In contrast, an LLM is typically trained on a \emph{broad swath of text} scraped from the Internet, predominantly in one language (most often English) with a mix of others, and not specifically on parallel data. For instance, GPT-style models are trained on web pages, books, and articles comprising many billions of words, of which only a small fraction may be parallel or even multilingual content \citep{blevins_language_2022, chowdhery_palm_2022}. As a result, NMT systems have in-domain strength for translation but limited “knowledge” outside language mappings, whereas LLMs boast vast “knowledge” and multilingual exposure but only \emph{implicit} translation mappings.

\subsection{Training Objective}

The dedicated NMT’s training objective is explicitly to maximize \emph{translation likelihood}, essentially \emph{directly training to translate}. The loss function compares the model’s output with reference translations for each source sentence. The LLM’s objective, by contrast, is \emph{next-word prediction} (language modeling) across its training texts. No direct signal tells the LLM “this is a translation of that.” Instead, at best, some training documents might include translations or multilingual sentence collections, and the LLM learns to predict those sequences. In short, translation is the \emph{end goal} for NMT but a mere \emph{side effect} for LLMs.

\subsection{Inference and Use}

Using an NMT model typically means feeding a source sentence into the model’s encoder and letting the decoder produce a translation. It is a straightforward, single-shot process. As already noted, using an LLM for translation requires crafting a suitable \emph{prompt}. For example, to translate from English to Spanish, one might prompt the LLM with: \texttt{\small{Translate the following English sentence to Spanish: <source sentence>}}. The LLM then generates the Spanish text as a continuation. The quality of the LLM’s translation can be sensitive to prompt wording \citep{jiao_is_2023, vilar_prompting_2023, peng_towards_2023}. The presence of examples (few-shot prompting; more on this below) or specifying the desired style can change results. By contrast, an NMT model’s output is largely deterministic given the source. This means that LLMs offer \emph{more flexibility} (one can prompt for a literal translation or a creative one, or ask an LLM to translate \emph{and explain}), but also \emph{more variability}. LLMs might ignore the instruction if not phrased clearly or might produce extra text (like apologies or analysis) if the prompt isn’t tight \citep{zhang_prompting_2023, peng_towards_2023, lyu_new_2023, balashov-etal-2025-translation}.

\subsection{Quality and Evaluation}\label{quality-and-evaluation}

How do these approaches compare in translation quality? Earlier  studies benchmarked LLMs against dedicated MT systems. For example, \citet{zhang_machine_2023} evaluated 15 open-source language models on translation tasks via prompting and compared to fine-tuned NMT models. They found that a bit of fine-tuning or domain adaptation (e.g. via QLoRA, \citealt{dettmers_qlora_2023}) on a small translation dataset boosts the translation performance of the model very significantly. A more extensive study showed that GPT-4 (which was a leading model at that time) outperformed Meta’s NLLB (No Language Left Behind, a specialized multilingual NMT model, \citealt{team_no_2022}) in about 40\% of evaluated translation directions, but still underperformed a strong commercial engine (Google Translate) on many low-resource language pairs \citep{zhu_multilingual_2024}. The consensus emerging from the work reported in 2023–2024 was that LLMs are remarkably good \emph{general translators}, especially for high-resource languages and with careful prompting, but a dedicated NMT that is tailored to a specific \emph{domain} or \emph{style} can still have an edge in its niche. For example, for technical texts with specialized terminology, a fine-tuned NMT might preserve terminology more consistently than an LLM, unless the LLM is explicitly guided or instructed \citep{hendy_how_2023}. But the field is developing very rapidly.\footnote{For a recent industry report, see \citealt{intento_inc_generative_2025}. For a recent linguistically-focused comparison of NMT, LLM, and human translation outputs, see \citealt{sizov-etal-2024-analysing}. \citealp{kong_evaluation_2025} is an example of a comparative user study of the translation performance of ChatGPT-4 and Google Translate in a very practical domain of hospital discharge instructions.}

\subsection{Multilingualism}

While some recent NMT models are multilingual by design (i.e. trained on parallel sentences in multiple languages and capable of translating between a chosen pair; e.g. NLLB), most are bilingual (one language pair). LLMs, in contrast, tend to be trained on multiple languages as a necessity: even if 90–95\% of the data is English, the remaining 5–10\% might cover dozens of languages \citep{blevins_language_2022, meta_llama_2025}. As a result, LLMs often have broader multilingual scope than a given NMT model \citep{cui_multilingual_2025}. A single LLM like GPT\nobreakdash-5 or Llama can translate between many languages (English to French, French to German, Japanese to Russian, etc.), whereas an NMT model is usually limited to the directions it was trained on (unless it’s a massively multilingual model like NLLB). This breadth is advantageous, but it comes with uncertainty: the LLM may not be equally strong in all languages. Some languages with very little representation in pretraining data see much lower translation quality.\footnote{E.g., \citet{zhu_multilingual_2024} note that GPT-4 still struggled on several African and Southeast Asian languages.} NMT models can also struggle with low-resource languages, of course, but one can explicitly gather parallel data for those languages to train or fine-tune the model, whereas an LLM’s fixed pretraining mix might under-represent them (unless augmented with further training).

\subsection{Behavior and Errors}

NMT and LLM systems also tend to exhibit different failure modes. NMT models, when they err, often produce mistranslations that are \emph{locally} implausible (e.g. garbled or nonsensical phrases) or they might drop segments of the input (omissions) if attention-based alignment fails. LLMs, by contrast, rarely produce outright garble; thanks to their strong language modeling priors their errors are usually fluent but can be \emph{misleading}. A common LLM translation error is \emph{hallucination}---inserting content absent from the source, or \emph{over-smoothing}---translating idiomatic expressions literally, thus losing nuance.\footnote{But \citet{he_exploring_2024} report that prompting an LLM to mimic a human translator’s step-by-step process can reduce such deficiencies.} LLMs might also be inconsistent in style or level of formality unless instructed, whereas an NMT system can be constrained by training data style or use explicit formality tags. These differences remind us that an LLM is trying to produce a \emph{plausible continuation} given the prompt and its training, whereas an NMT is laser-focused on reproducing the source content in target form.

\subsection{Specialization Versus Generalization}

In many cases, the contrast between NMT models and LLMs could be framed in terms of \emph{specialization} versus \emph{generalization}. NMT models tend to be \emph{specialists}---trained explicitly for translation, limited in scope and language directions but efficient and typically reliable within those limits. LLMs tend to be \emph{generalists}---not trained for translation in particular but endowed with a wide range of “knowledge” and linguistic background from which their translation abilities emerge. This generalist nature means LLMs can sometimes surprise with creative or context-aware translations (using their vast “knowledge” to resolve ambiguities that an NMT might be blind to), but it also means they might not always prioritize strict translation fidelity unless coaxed.

Understanding these differences is important as we examine how LLMs are being used for translation and other multilingual tasks in practice (Section \ref{recent-uses-of-LLMs}), how they \emph{generate} translation internally (Section \ref{how-llms-translate}), and \emph{why} they perform so well despite the lack of direct translation training (Section \ref{what-explains}). The next section surveys the rapidly growing array of applications and experiments with LLMs in translation-related tasks, setting the stage for deeper analysis of their abilities.

\section{Recent Uses of LLMs in Translation Tasks} \label{recent-uses-of-LLMs}

The application of LLMs to translation-related tasks began almost as soon as LLMs themselves burst onto the scene \citep{brown_language_2020, chowdhery_palm_2022, openai_gpt-4_2023}. Researchers and practitioners have tested cutting-edge LLMs on various multilingual tasks, often with impressive results. In this section, I review recent uses of LLMs in translation and other cross-lingual contexts. Key themes include zero-shot vs. few-shot prompting, performance on low-resource languages, pivot translation strategies, chain-of-thought prompting for translation, and exploiting the inherently multilingual nature of latent representations in LLMs.

\subsection{Zero- and Few-Shot Translation with LLMs}\label{zero-few-shot}

One of the headline capabilities of LLMs is zero-shot learning---performing a task without direct task-specific training \citep{brown_language_2020}. Translation has become a primary example. Prompting an LLM with a simple instruction “\texttt{\small Translate the following sentence to Mandarin:\_\_\_\_\_\_\_\_}” can yield a credible translation even if the model was never explicitly trained on parallel data for that language. Early tests with GPT-3 showed surprisingly competent zero-shot translation for several languages \citep{castilho_online_2023}, sparking immediate interest in using such models as universal translators \citep{savenkov_gpt-3_2023}.

Subsequent research has systematically evaluated zero-shot vs. few-shot prompting for translation (e.g. \citealt
{zhang_machine_2023}). Few-shot prompting means providing a few example translations in the prompt before asking the model to translate a new sentence. For instance, one might input:

\begin{displayquote}
\texttt{\small \textcolor{blue}{English: Hello, how are you? $\rightarrow$ \textcolor{red}{Spanish: Hola, ¿cómo estás?};\linebreak English: My name is John. $\rightarrow$ \textcolor{red}{Spanish: Me llamo John};\linebreak English: Where is the library? $\rightarrow$ \textcolor{red}{Spanish:}}}
\end{displayquote}

\noindent and let the model continue. This technique often boosts translation accuracy, especially for models not specifically instruction-tuned. \citet{garcia_unreasonable_2023} dubbed it the “unreasonable effectiveness of few-shot learning,” finding that even 1–2 examples can significantly improve the translation output output for certain LLMs. The improvement is typically more pronounced for smaller or less instruction-savvy models. On the other hand, for GPT-4 and later models, the gap between zero-shot and few-shot regimes is smaller, since the former is already strong. Later work noted that few-shot translation prompting can, in effect, make up for the lack of translation-specific instruction tuning, or of any instruction tuning.

In fact, curious phenomena have been observed. For example, \citet{zhu_multilingual_2024} found that, when given in-context exemplars, LLMs sometimes ignore “unreasonable instructions” asking the model to translate into a different language or do something else altogether, e.g. summarize a sentence instead of translating it. The authors also found that \emph{cross-lingual} exemplars (in different language pairs than the target pair) can sometimes provide better guidance for low-resource translation than same-pair exemplars. Thus, inserting several RU-EN sentence pairs in the prompt, following by un unrelated Chinese sentence produces a better English translation of it than ZH-EN examples. A hypothetical explanation may be that showing the model examples of, say, RU-EN translation may activate its “translation mode” generally, even if we actually want it to translate from Chinese to English, a direction for which it saw no direct examples. This aligns with the idea that the LLM has a general concept of “translating” that is somewhat language-agnostic. The general lesson appears to be that instruction semantics can be overridden by example context.

Both zero- and few-shot regimes raise intriguing questions about \emph{how} and \emph{where} translation happens in a fully trained LLM. I will discuss them in Section \ref{how-llms-translate}.

\subsection{Low-Resource Languages}\label{low-resource-languages}

As noted, LLMs’ translation performance appears to mirror global data inequalities: it tends to be strong for high-resource languages and weaker for low-resource ones. Pivoting through English (Section \ref{pivoting-intermadiate}) is sometimes used to deal with this challenge.

For extremely low-resource languages, with virtually no representation in the LLM’s pre-training data, neither pivoting nor prompt optimization may suffice. In such cases, researchers have explored fine-tuning LLMs on supplemental parallel data in a low-resource language pair. The hope is the LLM’s general knowledge transfers, and only minimal data is needed to teach it a new language mapping. Early results show some promise, but the cost of fine-tuning an LLM can be prohibitive. Fortunately, techniques such as LoRA (Low-Rank Adaptation, \citealt{hu_lora_2021}) can be used to inject translation capabilities for new languages into an existing LLM at relatively low compute cost. Thus \citet{zhang_machine_2023} used an even more efficient method (QLoRA, \citealt{dettmers_qlora_2023}) to effectively turn a general LLM into a specialized translator with just a fraction of parameters updated, something that can be done on a single consumer-grade GPU.

\subsection{Pivoting and Intermediate Languages}\label{pivoting-intermadiate}

Pivoting (translating through an intermediate language) has been a longstanding strategy in MT for dealing with language pairs lacking direct parallel data. With LLMs, explicit pivoting can be attempted by prompting the model to go through two steps; for example: “\texttt{\small{Translate this sentence from German to English first; then translate the \linebreak English output to Hindi.}}” This prompt can sometimes yield a better Hindi translation than prompting the model to translate directly from German to Hindi (cf. \citealt{jiao_is_2023}).

\subsection{Chain-of-Thought Prompting for Translation}\label{cot-prompting-for-translation}

Chain-of-thought (CoT) prompting is a technique where the model is asked to produce intermediate reasoning steps before the final answer \citep{wei_chain--thought_2022}. Originally developed for math and logic problems, it has been applied to help LLMs improve their translation output by following, in some cases, a human translation workflow model. For example, \citet{briakou_translating_2024} engaged Gemini 1.5 Pro in a multi-turn interaction involving pre-translation research, drafting, refining, and proofreading, which resulted in notable quality improvements.

CoT prompting could be particularly useful when dealing with longer documents. \citet{he_exploring_2024} adopted this approach and asked LLMs to generate keywords, topics, and potential translations of tricky terms before attempting the full translation and then used those to guide the process. Since state-of-the-art LLMs have very large context windows (1M+ tokens in some cases), one can feed an entire document and prompt the model to work section by section. Researchers are actively exploring this: for instance, how to prompt an LLM to decide when to translate literally and when to adapt, essentially giving the model a strategy akin to a human translator’s decision-making \citep{jiao_parrot_2023, lu2024mqmapehighqualityerrorannotation}. This ventures into  the territory of interactive translation with LLMs: rather than using a single prompt, one can craft a sequence of prompts for a genuinely collaborative scenario. There have been experiments using LLMs as translator assistants, where the human can query the model for alternate phrasings, explanations of source text nuances, error annotation, and more \citep{jiao_parrot_2023, ki_guiding_2024, treviso_xtower_2024}.

Another interesting recent study develops “compositional translation,” a new method that uses LLMs to improve translation quality for low-resource languages \citep{zebaze_compositional_2025}. The approach works by breaking down complex sentences into simpler, semantically coherent phrases, translating each phrase using contextually similar examples, and then merging these translations into a coherent final result. Experiments show that this method can outperform traditional few-shot translation techniques in LLMs, making it a promising solution for low-resource challenges.

Finally, explicit pivoting through English (Section \ref{pivoting-intermadiate}) is another, less sophisticated case of CoT prompting.

\subsection{Multilingual Latent Representations and Cross-Lingual Tasks}

Beyond direct translation, LLMs have shown capability in a variety of \emph{cross-lingual tasks}, such as cross-lingual question answering (answering in English from a French context, etc.), multilingual summarization (summarizing a document in another language), code-switched dialogue understanding, and so on \citep{chua_crosslingual_2023, gao_understanding_2025, blum2025rosettastoneunificationforces}. Presumably, all these tasks leverage the model’s \emph{multilingual latent space}. But what, exactly, is the shape of this space \citep{ravisankar_can_2025}? Where in the model should we look for it? And what methods could be used to study it?

The next section reviews key recent work addressing these questions, offering additional background and a valuable framework for the investigation of the origins of LLMs’ translation abilities in their training data in Section \ref{what-explains}.

\section{How Do LLMs Translate? Where Does Translation Happen in Them?} \label{how-llms-translate}

\subsection{In-Context Translation}\label{in-context-translation}

Few-shot translation mentioned above (Section \ref{zero-few-shot}) is a remarkable example of in-context learning in LLMs \citep{brown_language_2020, dong_survey_2024}. The “unreasonable effectiveness” \citep{garcia_unreasonable_2023} of in-context translation and its ability to override semantically inadequate prompt instructions \citep{zhu_multilingual_2024} raise questions about its underlying computational mechanisms.

\citet{sia_where_2024} approached these questions by asking \emph{at what stage} in the computation LLMs recognize and execute the translation task in a few-shot mode. By using a technique called “layer-from context masking,” the authors selectively removed the model’s ability to attend to prompt instructions and examples from specific layers onward. This method allowed them to locate a “task recognition point”---around layer 14 of 32 in Llama 7B, for instance---where the model has internalized the task and no longer needs to consult the context to generate accurate translations. Before this point, masking the context significantly degraded performance; after it, performance remained stable even when the context was masked. This suggests a three-phase process: an initial stage with minimal context influence, a middle stage of active task location, and a final stage where execution proceeds independently of the prompt.

Further experiments identified these task recognition layers as “critical layers.” Masking them severely impacted performance, whereas masking later layers had minimal effect, indicating redundancy. Moreover, fine-tuning experiments using lightweight LoRA adapters showed the most gains when applied to these middle layers, underscoring their central role in learning the translation task.

\subsection{Cross-Lingual Representation Alignment}\label{x-ling-representation-alignment}

An intriguing earlier finding by \citet{wen-yi_hyperpolyglot_2023} was that some language models cluster semantically similar words from different languages together at the \emph{very first}, often ignored, \emph{embedding} layer. Specifically, the authors found that, without explicit bilingual training, the token embeddings of mT5-XL \citep{xue_mt5_2021} often align across languages for concepts; for example, the vectors for \emph{når} (“when,” Danish), \emph{cuando} (“when,” Spanish), \emph{quando} (“when,” Portuguese), \<k/s> (“when,” Hebrew), and \emph{quand} (“when,” French) are all close neighbors. A similar phenomenon was earlier observed in models like mBERT \citep{devlin_bert_2019}, but not in the input embedding layers.

It should be noted that mT5 is an \emph{encoder-decoder} model trained on a balanced mix of 100+ languages with the “masked span prediction” training objective, and designed for text-to-text tasks, including translation, summarization, and so on. mBERT and its descendants such as XLM-R, on the other hand, are \emph{encoder-only} multilingual models similar to BERT but trained for multilingual understanding tasks such as classification or NER, and not for generation.

What about pristine \emph{decoder-only} models? Are they also capable of learning cross-lingual representation alignment and cross-lingual transfer during pre-training? \citet{hua_mothello_2024} approach these questions with a controlled training of GPT-2 models on a multilingual extension of a toy synthetic language that was earlier used to model legal moves in the Othello board game \citep{li_emergent_2024}. While limited both in scope and scale, their findings suggest that relatively small decoder-only models do not learn a common representation across multiple artificial languages when the models are exposed to them separately. However, adding “anchor tokens”---lexical items that are shared across languages---to the training sequences improves cross-lingual representation alignment quite considerably. As the authors note, mOthello languages with their simplistic grammar and only 180 tokens are a far cry from natural languages. Yet the catalyzing role of “anchor tokens” in “Global” (Section \ref{intro}) multilingual representation learning is notable and consistent with the role of shared subword vocabularies and “seed lexicons” earlier studied in other settings, such as unsupervised MT \citep{artetxe_unsupervised_2018, sogaard_limitations_2018, conneau_unsupervised_2020} and multilingual word embeddings more generally \citep[Ch. 12]{koehn_neural_2020}.

Experiments with toy models, synthetic data, and narrowly scoped tasks offer a distinctive advantage: they give researchers fine-grained control over training dynamics. In a recent study, \citet{blum2025rosettastoneunificationforces} trained tiny 2M-parameter transformers from scratch on synthetic multilingual corpora and uncovered an early learning phase during which models either unify or segregate the same facts across languages. Only the unification regime yields robust cross-lingual transfer. The authors show that the strength of unification closely tracks cross-lingual factual accuracy and can be causally modulated through dataset design. In particular, when language identity is easy to infer and predictive of simple labels, models drift toward language-siloed representations; by contrast, balancing the data and attenuating language-specific surface cues encourages cross-lingual unification.

\subsection{Do LLMs Pivot Through English on Their Own?}\label{do-llms-pivot-on-their-own}

Already mentioned in Section \ref{pivoting-intermadiate}, pivoting through an intermediate language (most often English) has been widely exploited in the MT industry for a long time. Recent user studies have extended this approach to LLMs by explicitly asking the models to go through two stages, and sometimes found improved accuracy.

\begin{figure}[h]
    \centering
    \includegraphics[width=0.75\textwidth]{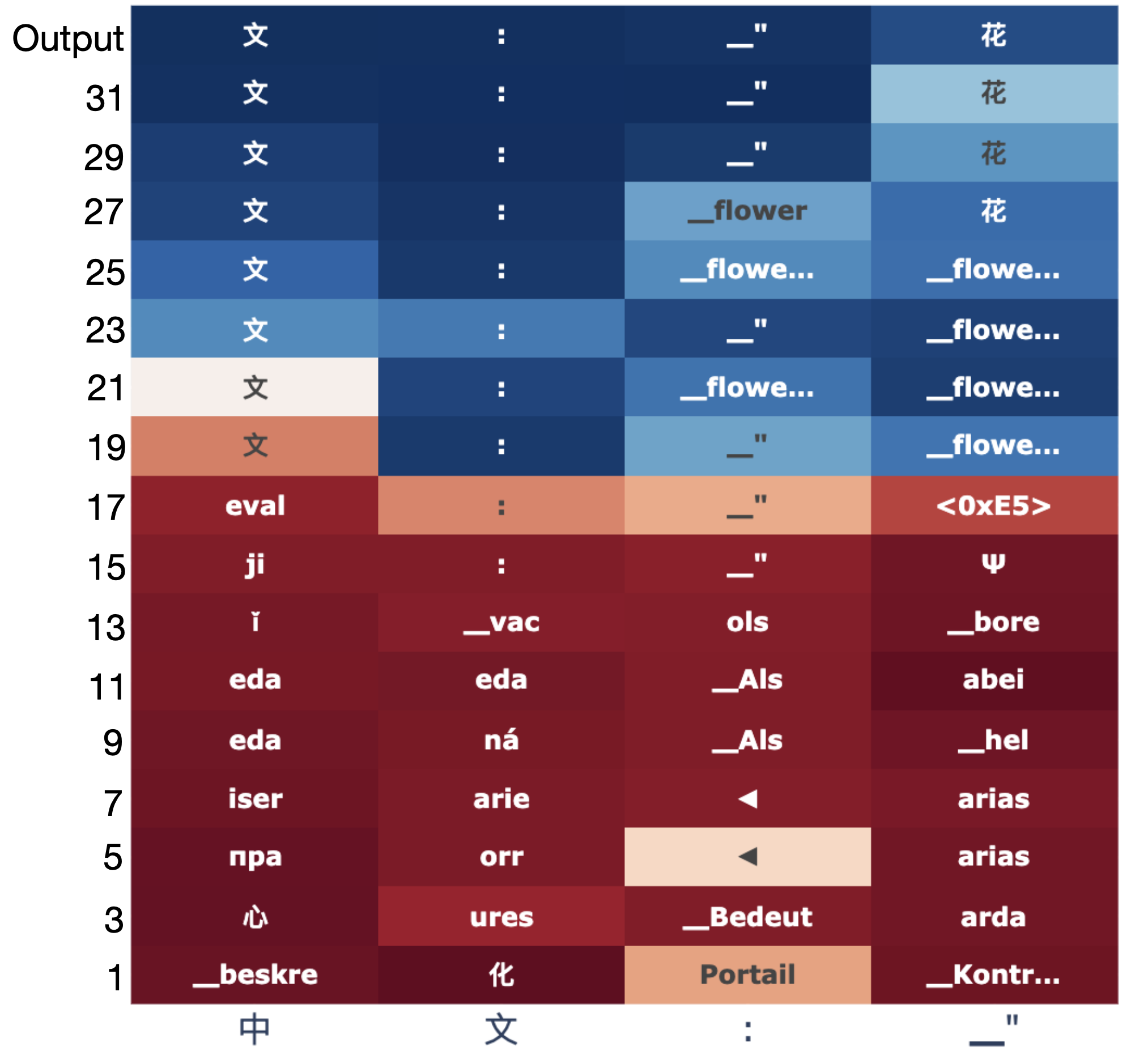}
    \caption{Llama-2 7B translates ‘fleur’ to ‘\begin{CJK*}{UTF8}{bsmi}花\end{CJK*}’ by seemingly pivoting through ‘flower’ in intermediate layers. Figure~1 in \citet{wendler_llamas_2024}.}
    \label{fig:figure4}
\end{figure}

Interestingly, evidence suggests LLMs themselves might internally be doing something like this. \citet{wendler_llamas_2024} provided insight by asking: do multilingual LLMs use English as an internal pivot language? They prompted Llama-2 models to translate a single French word ‘fleur’ to Chinese with a few-shot prompt (where ‘\begin{CJK*}{UTF8}{bsmi}中文\end{CJK*}’ means “Chinese”):

\begin{displayquote}

\small \textcolor{blue}{Francais: "vertu" -} \textcolor{red}{\begin{CJK*}{UTF8}{bsmi}中文\end{CJK*}: "\begin{CJK*}{UTF8}{bsmi}德\end{CJK*}"}

\textcolor{blue}{Francais: "siege" -} \textcolor{red}{\begin{CJK*}{UTF8}{bsmi}中文\end{CJK*}: "\begin{CJK*}{UTF8}{bsmi}座\end{CJK*}"}

\textcolor{blue}{Francais: "neige" -} \textcolor{red}{\begin{CJK*}{UTF8}{bsmi}中文\end{CJK*}: "\begin{CJK*}{UTF8}{bsmi}雪\end{CJK*}"}

\textcolor{blue}{Francais: "montagne" -} \textcolor{red}{\begin{CJK*}{UTF8}{bsmi}中文\end{CJK*}: "\begin{CJK*}{UTF8}{bsmi}山\end{CJK*}"}

\textcolor{blue}{Francais: "fleur" -} \textcolor{red}{\begin{CJK*}{UTF8}{bsmi}中文\end{CJK*}: "}

\end{displayquote}

\noindent and applied a technique called \emph{logit lens} \citep{nostalgebraist_interpreting_2020} to unembed models’ intermediate layers early to see what they encode. They found that the model’s representation of the French input drifts closer to English semantic space before settling into the target language space (i.e. Chinese) (Figure~\ref{fig:figure4}).

In essence, an LLM might internally translate everything to a kind of “English-centric interlingua” as a steppingstone. If so, then explicitly pivoting via English might resonate with the model’s inner workings. Indeed, some prompting approaches encourage the model to think aloud in English before producing the final translation in the target language. And more recent “reasoning” models sometimes do it on their own \citep{liu_new_2025, chen_evaluating_2025}, using English as a kind of “metalanguage.”\footnote{Unfortunately, this interesting metalinguistic phenomenon lies outside the scope of the paper. Could it form the basis of another distinct mode of translation as "extended inference" analogous to similar operations sometimes performed by human translators when they encounter difficult cases? This question is definitely worth exploring. Thanks to Raphaël Millière for drawing my attention to it.}

\subsection{Language-Agnostic Representation?}\label{language-agnostic-representation}

\begin{figure}[H]
    \centering
    \includegraphics[width=1.0\textwidth]{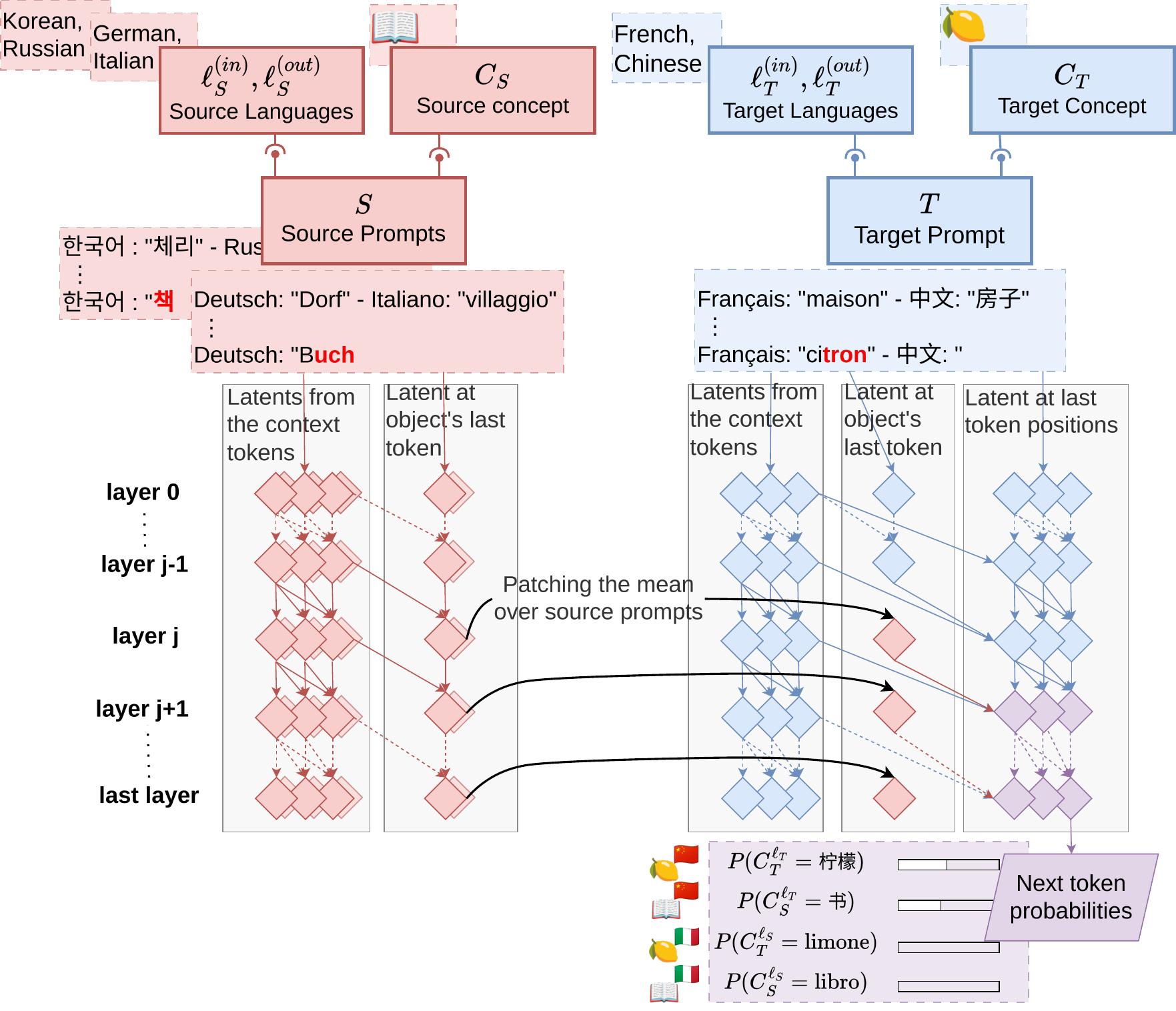}
    \caption{Figure~1 in \citealt{dumas_separating_2025}. A hidden state from the source prompt (red, left) at a certain layer is inserted into the forward pass of the target prompt (blue, right) at the corresponding layer. The source prompt here provides an example translation (German ‘Dorf’ $\rightarrow$ Italian ‘villaggio’) and then asks to translate ‘Buch’ (\emph{book}) to Italian, while the target prompt asks to translate French ‘citron’ (\emph{lemon}) to Chinese. By patching the red latent into the blue context at different layers, the researchers observed changes in the predicted translation. Early-layer patching yields the correct target concept in the target language (e.g. \mbox{‘\zh{柠檬}’} for \emph{lemon} in Chinese), mid-layer patching yields the correct concept but in the wrong language (‘limone’ = \emph{lemon} in Italian), and late-layer patching yields the wrong concept in the source language (‘libro’ = \emph{book} in Italian).        
      \label{fig:figure5}}
\end{figure}

But do LLMs actually “think” in English when performing multilingual tasks \citep{schut_multilingual_2025}? Or do they “think” in a truly interlingual way, with English emerging in the intermediate layers simply because it happens to dominate the training data \citep{wang-etal-2025-lost-multilinguality}? Wendler et al.’s early decoding study \citep{wendler_llamas_2024, lim2025languagespecificlatentprocesshinders} did not allow them to address this question head-on. \citet{dumas_separating_2025} take up this challenge with a more nuanced mechanistic interpretability approach based on \emph{activation patching}. They designed simple translation prompts in different languages to serve as two contexts: a “source prompt” (translating a word from, say, German to Italian) and a “target prompt” (translating a \emph{different} word from, say, French to Chinese). The authors ran Llama-2 7B on both prompts and then \emph{swapped} part of the internal state between them mid-process. More precisely, at a chosen transformer layer, they took the hidden latent representation of the last token from the source prompt’s forward pass and injected it into the target prompt’s forward pass at the same layer. By observing how this surgical insertion alters the model’s next-word prediction, they can tell what information that latent was carrying (Figure~\ref{fig:figure5}).

The outcome was quite striking: different layers separated the handling of \emph{language} and \emph{meaning}. When \citet{dumas_separating_2025} patched at \emph{early} layers (closer to the input), Llama-2 still produced the correct word in the target language, as if nothing had changed. But patching at a \emph{middle} layer caused a curious swap: the model generated the \emph{correct concept} but in the \emph{wrong language} (specifically, it used the source prompt’s language). And patching at a \emph{late} layer flipped both aspects---the concept as well as the language, effectively following the source prompt’s content entirely. This layered behavior reveals that Llama-2 selects the output language earlier in the network, and determines the actual concept to translate later, in deeper layers. In other words, the model first decides “I need to respond in Chinese” (target language) before it figures out “what is a lemon?” in the context of translation.

To dig deeper, Dumas and co-authors formulated two competing hypotheses about Llama-2’s internal mechanism: H1 (“Disentangled Representation”): the model keeps language and meaning separate, computing a language-agnostic concept first and then translating it into the required language; and H2 (“Entangled Representation”): the model’s concept representation is always tied to a particular language, so it directly converts from a source-language word to a target-language word in one entwined step. The layer patching results hint at H1: one could swap languages while keeping the same concept, and vice versa. To confirm this, the authors \emph{averaged} the latent concept representations for a given word across \emph{multiple} source languages and used this mean representation in a translation prompt. Remarkably, this language-agnostic average did not confuse the model or degrade its performance; in fact, the performance was improved! In other words, forcing the model to translate from a blended, language-neutral hidden state yielded even better results. This counterintuitive outcome adds force to the claim that the model’s internal representation of a concept may be genuinely “language-neutral.”

The authors note the limitations of their study which operates with single word prompts and very simple unambiguous concepts, explicitly acknowledging that richer lexical items, multi‑word expressions, sentential context, discourse phenomena and language‑specific idioms remain outside the scope of this approach. It is unclear whether the demonstrated disentanglement of language and meaning carries over to full-sentence translation. All the reported experiments use open-source models (in addition to Llama‑2 7B, the authors ran experiments with Llama‑3 8B, Mistral 7B, Qwen 7B, Aya 8B, Gemma 2B). While the authors replicated some of their findings on a 70B Llama, they obviously could not test frontier‑scale, instruction‑tuned or RLHF‑aligned systems. Differences in the model size, training mix, and alignment objectives may alter how larger “consumer‑grade” LLMs partition language and meaning. Finally, while activation patching is a powerful tool it may introduce artifacts and does not, by itself, guarantee that the model would naturally traverse the identified internal states during ordinary inference. But despite these limitations, \mbox{Dumas} et al.’s work contributes important insights to the ongoing debate about the nature of multilingual representation in LLMs.

A further step in this direction was recently undertaken by Anthropic’s transformer circuits team \citep{lindsey_biology_2025} who used a newly developed tool called \emph{attribution graphs} to trace how certain features (interpretable patterns of activation in the network) interact to produce the model’s output.

An attribution graph is essentially a causal map of the model’s internal computation for a given prompt, where the nodes are interpretable features and the directed edges show how activating one feature leads to effects on others and ultimately the output. Attribution graphs trace how the model’s internal components contribute to a specific answer, much like an information flowchart. A feature, in this context, means an interpretable direction in the model’s activation space---a pattern of neural activity that consistently represents a particular concept. Features can range from low-level patterns such as detecting a specific token or character to high-level semantic themes such as a concept of size, negation, or French language. By identifying these features and how they connect with each other in an attribution graph, Anthropic’s researchers formulated hypotheses about what sub-computations the model is performing to generate an output. Crucially, this approach yields an interpretable pattern of the model’s internals as a candidate explanation of which concepts the model used and how they interacted to produce a given response. The authors then validated these explanations with \emph{intervention} experiments by actively editing features in the model’s computations to test whether the predicted causal roles hold up. This combination of attribution graphs (visualizing which features matter) and feature-level interventions (probing what happens if we change them) serves as a powerful tool for mechanistic interpretability.

In a bit more detail, the tool was put to work in a simple translation-like task: the prompt “\texttt{The opposite of ‘small’ is …}” posed in three different languages (English, French, and Chinese). Despite the surface language differences, all three prompts have the same meaning, and their correct completions are semantically equivalent: ‘big’, ‘grand’, and ‘\begin{CJK*}{UTF8}{bsmi}大\end{CJK*}’. The question is: does the model (Claude 3.5 Haiku) respond to these prompts using one language-neutral reasoning process, or does it handle each language separately? The attribution graphs reveal a striking answer: the model uses very \emph{similar internal circuits} for all three languages combining shared cross-lingual features with language-specific ones. In each case, Claude first activates a set of language-independent features that recognize the task of finding the antonym of ‘small’. These include an abstract concept of “smallness” and an “opposite-of” operation feature. Once this conceptual step finds the idea of \emph{big} (the opposite of \emph{small}), the model then engages language-specific features to express that idea in the appropriate tongue. In other words, Claude appears to “think” in a language-neutral way about the meaning (identifying the “opposite of small” as the concept \emph{big}), and only then translate that concept into the word forms ‘big’, ‘grand’, or ‘\begin{CJK*}{UTF8}{bsmi}大\end{CJK*}’ depending on whether the context is English, French, or Chinese. There are features that act as language markers; for example, an \texttt{\textcolor{blue}{open-quote-in-French}} feature or a \texttt{\textcolor{blue}{Chinese-context}} feature, which ensure the answer is framed in the correct language with the right quotation marks. But the core semantic features (for \emph{small} and its opposite) are multilingual: the model has learned a single abstract notion of \emph{small} that spans languages, rather than distinct unrelated notions for each language. The authors describe the overall computation as having three factorable parts: \emph{Operation} (finding an antonym), \emph{Operand} (the concept of \emph{small}), and \emph{Language} (which language’s vocabulary to use for the answer). This decomposition suggests that the model might indeed be employing something like an internal “interlingua” for meaning, combined with a separate layer for language-specific expression.

\begin{figure}[H]
    \centering
    \includegraphics[width=1.0\textwidth]{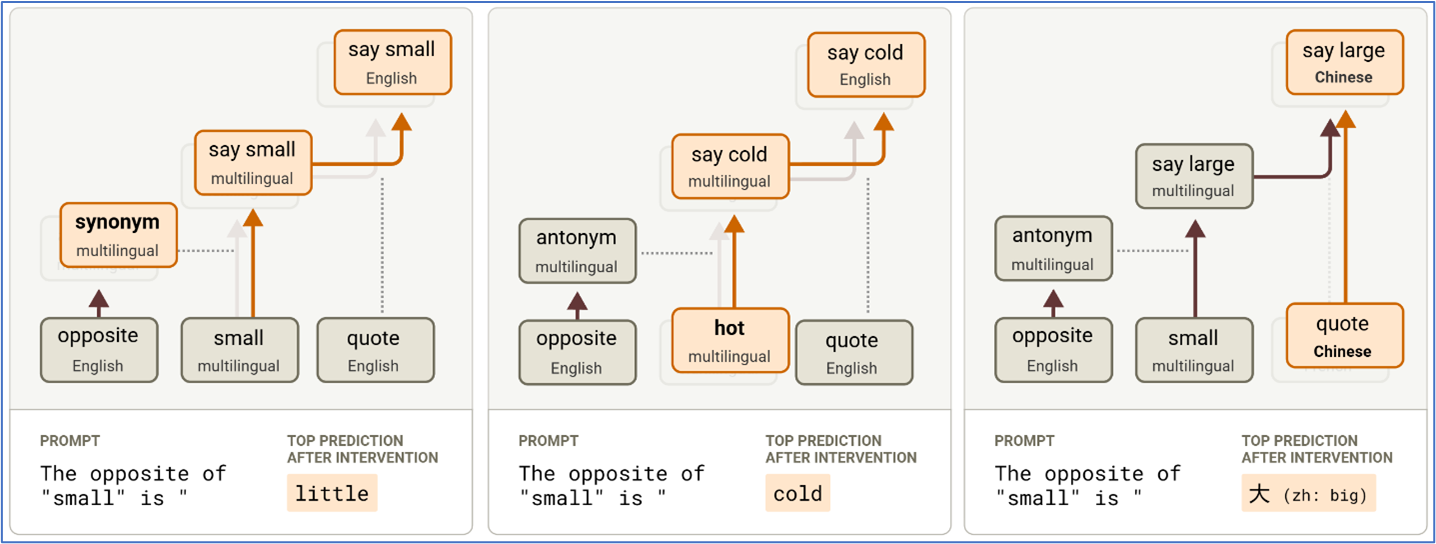}
    \caption{Three targeted interventions on Claude’s multilingual circuit (\citealt{lindsey_biology_2025}, Figure~18). Left: \emph{Operation Swap}---replacing the antonym operation features with synonym features (while the input still asks for “opposite”) causes the model to output ‘little’ (a synonym of ‘small’) instead of the antonym ‘big’. This shows that toggling the internal operation from “find opposite” to “find similar” yields a corresponding change in output. Center: \emph{Operand Swap}---replacing the features encoding the concept \emph{small} with those for \emph{hot} causes the model to respond with ‘cold’ (the antonym of ‘hot’). The prompt remains “The opposite of ‘small’ is…”, but internally the model now treats ‘small’ as if it were ‘hot’, demonstrating that the content of the adjective can be swapped independently of the prompt’s wording. Right: \emph{Language Swap}---replacing English-language context features with Chinese ones makes the model output ‘\begin{CJK*}{UTF8}{bsmi}大\end{CJK*}’ (Chinese for ‘big’) for the English prompt. The model still found the concept \emph{big} as the opposite of \emph{small} but expressed it in a different language. In each case, only one aspect of the computation is changed (function, content, or language), and the model’s behavior changes in the predicted way, confirming that these internal features indeed govern those distinct components of the translation task.}
    \label{fig:figure6}
\end{figure}

To confirm that these parts (\emph{Operation}, \emph{Operand}, \emph{Language}) are truly separable in Claude’s internal representation, the authors performed three kinds of causal intervention experiments, each targeting one part while leaving the others unchanged. Figure~\ref{fig:figure6} illustrates these three interventions and their effects. In each case, the researchers locate the relevant features in the attribution graph and surgically swap or alter them mid-computation to see how the model’s output changes, thereby testing whether that facet of the circuit is indeed functioning as hypothesized.

The results confirmed the expectation that Claude’s internal processing has a “modular” structure: one module figures out the task (antonym vs. synonym), another handles the core meaning of the words involved, and a third handles the language context. Each can be perturbed on its own. This level of interpretability is quite remarkable: it is as if one opened the model’s “brain” and identified a sub-network for “opposite meaning,” a sub-network for “the concept of \emph{small},” and a sub-network for “speaking French,” and demonstrated each one’s role by poking it. This offers a fascinating parallel to the idea that meanings are not intrinsically tied to words: the model  \emph{generates} the appropriate word when needed, implying it has an independent grasp of the meaning itself. Furthermore, the model’s ability to swap out  \emph{Operation} and  \emph{Operand} independently points to a form of  \emph{compositional representation} inside the neural network. The classical hallmark of \emph{compositionality} in linguistic semantics is the requirement that “the meaning of an expression [be] a function of the meanings of its parts and of the way they are syntactically combined” \citep[281]{partee_barbara_compositionality_1984}. Here we see it at work: the model’s internal process for “the opposite of X” can be decomposed into a part that understands the  \emph{relation} “opposite-of” and a part that provides the  \emph{content} “X,” and these parts can be intervened on separately. The “antonym-finding” operation acts almost like a built-in function that can be applied to different inputs (‘small’ or ‘hot’ or others) to yield different outputs. Importantly, the concept of  \emph{small} is represented in a way that is not entangled with the fact that it was asked in French or English; it is a “portable” piece of information.

On a cautionary note, Anthropic’s team also observed that English seems to be the model’s “default” language in this internal process; which is hardly surprising. They describe English as “mechanistically privileged,” meaning that, in the model’s circuits, English has a built-in advantage or priority. How does this show up mechanistically? It turns out that those abstract “say \emph{big}” features (the ones that correspond to expressing the concept of \emph{big}) have a stronger direct influence on producing the English word ‘big’ than the equivalents in French or Chinese. In Claude’s neural wiring, the path from the concept of \emph{big} to the English token ‘big’ is somewhat more straightforward and higher-weighted than the path to ‘grand’ or ‘\begin{CJK*}{UTF8}{bsmi}大\end{CJK*}’. For non-English outputs, the model relies a bit more on the extra “language context” features to get to the right word. Another subtle observation was that some English-specific features engage in what the authors call a “double inhibitory effect”: essentially, certain English features suppress competing features that would otherwise produce the English word when the context is not English, thus allowing the non-English word to come out. This complicated dance of inhibition and activation paints a picture in which the model retains a universal semantic representation (of \emph{big}) that is biased to map to English unless additional mechanisms redirect it. In more philosophical terms, Claude 3.5 has a kind of universal “mentalese” for concepts \citep{Fodor1975-FODTLO}, but it is skewed towards outputting English by default. Only when the model clearly detects a non-English context (quotes, characters, etc.) do these language-specific signals intervene to translate that internal thought into, say, French. This finding echoes the tension noted in the recent literature: some prior work had found evidence of a shared multilingual representation \citep{zhang_same_2024}, while others noted an English-centric bias \citep{schut_multilingual_2025}. Here, both appear to be true: the representations are genuinely multilingual, \emph{and} English is a privileged default internally. One could say the model has a \emph{universal} “mental dictionary” but with English entries written \textbf{in bold}.

All of that is still quite a distance away from real-life translation “in the wild.” And the indirect nature of the attribution graphs method, which operates on a separate "replacement model" rather than the original model, contributes to the limitations of this study. Nonetheless, cutting-edge mechanistic interpretability research casts considerable light on both questions stated in the title of this section: “How do LLMs translate?” and “Where does translation happen in them?” The answers are nuanced, depending on multiple factors involved in various experiments: zero- vs. few-shot prompting, single concepts (\emph{lemon}, \emph{book}, etc.) vs. concepts and functions (“the opposite of\_\_\_”, “a synonym of\_\_\_”), as well as model size and other details. But the evidence strongly suggests that LLMs, large and small, develop some kind of shared multilingual or language-neutral representations during training and then leverage it at inference time when prompted for translation-related tasks.

How do LLMs develop these representations in the first place? The answer lies in the models’ \emph{training data} and \emph{training dynamics}.

\section{What Explains the Origin of LLMs’ Translation Abilities?} \label{what-explains}

\subsection{The Minimal Role of Instruction Tuning}\label{minimal-role-of--instruction-tuning}

After their initial pre-training LLMs normally undergo supervised instruction tuning which improves their responsiveness to specific requests. This is done on a mixture of mostly proprietary data, but there is little doubt that translation tasks are included in the mix. In any case, they are present in open-source instruction tuning datasets such as Aya \citep{singh_aya_2024}. How important are they to the emergence of LLMs’ translation abilities?

Early experiments revealed rather erratic behavior of even the largest language models and their high sensitivity to the prompting details. For example, in response to the following prompt:

\begin{displayquote}

\texttt{\small{Translate French to English: Mon corps est un transformateur de soi, mais aussi un transformateur pour cette cire de langage.}}
\end{displayquote}

\noindent GPT-3 continued with another French sentence instead of translating \citep{reynolds_prompt_2021}.

Instruction tuning is partly about teaching the model how to respond in a way that humans find coherent and helpful. When a user asks, “\texttt{Translate from French to English},” the model leverages the multilingual knowledge it \emph{has already formed} during pretraining, but now it has learned a clearer instruction on how to \emph{apply} that knowledge (i.e., produce a direct translation rather than continuing in the original language).

Although there is still some dependence on the prompting details \citep{hendy_how_2023, peng_towards_2023, sizov-etal-2024-analysing}, in the latest LLMs suboptimal prompts tend to yield overly \emph{generous} rather than incorrect responses to translation tasks. Some models need to be \emph{restrained} with longer prompts in order to limit their output to what is needed (e.g. \citealt{balashov-etal-2025-translation}).

To be sure, decidedly vague or ambiguous prompts may still push a model in a wrong direction. And, as already noted, including translation examples in the prompt tends to override inadequate instructions. But there is every reason to believe that LLMs’ translation abilities are already there in a dormant state and are simply waiting to be unlocked with a good prompt.

So: while it may be tempting to attribute LLMs’ translation capabilities to explicit instruction tuning, it likely plays only a minor or supplementary role. Indeed, the translation capabilities of LLMs were already prominent in earlier, non-instruction-tuned models, such as GPT-2, GPT-3, and BLOOM. And they are stronger in non-instruction-tuned versions of more balanced multilingual models, such as Llama-3 and Llama-4, even if the latter are smaller in size. Due to their exposure to massive multilingual texts, LLMs inadvertently learn cross-lingual patterns and semantic mappings. Instruction tuning may enhance model responsiveness and specificity, but it does not fundamentally generate their cross-lingual capabilities. The end result is that, once prompted to translate, such a model can search its deeply learned space of multilingual representations and produce coherent translations, even for language pairs it has not been explicitly trained to translate in a supervised manner. Good prompts are more important than instruction tuning on translation examples.

At their core, LLMs develop their translation abilities during the pre-training phase. How does it happen?

\subsection{Incidental Bilingualism in Training Data}\label{incidental-bl}

In an earlier work, Eleftheria Briakou and colleagues \citep{briakou_searching_2023} approached this question by searching for “bilingual needles” in a haystack of PaLM’s \citep{chowdhery_palm_2022} training data. The authors hypothesized that \emph{incidental bilingualism}---the unintentional inclusion of bilingual text within single training instances---is a key source of PaLM’s translation capabilities. In other words, the 540B-parameter model may have learned to translate from naturally occurring translations and mixed-language content in its massive training corpus (780 billion tokens of crawled text from web pages, social media, books, and so on), even though it was never given dedicated parallel data. The idea is that, while combing through its training corpus, PaLM may have stumbled upon numerous “needles in the haystack”: sentences and documents that appear side by side in different languages, essentially acting as implicit translation examples. These bilingual snippets were not explicitly labeled as “training examples” for translation; rather, they were \emph{incidental}, arising naturally in the wild text. For example, a website might list a paragraph in English followed by the same paragraph in Spanish, or a forum post might include a quote in French with an English gloss. Briakou et al. suspected that such incidental bilingual data could be the primary source from which PaLM picked up its translation skills.

To test this hypothesis, the authors adopted a multi-step mixed-method approach, combining large-scale quantitative analysis with targeted qualitative inspection. The first step in the process was distinguishing bilingual from monolingual text in PaLM’s 2048-token-long training instances. In turns out that only about 1.4\% of them are bi- or multilingual, i.e. contain more than one language, and only 0.34\% contain at least one translated sentence pair: a sentence in one language directly paired with its translation in another. However, given the enormous scale of the corpus, this still amounts to a very large absolute volume of parallel data. In total, PaLM was unintentionally exposed to over 30 million translated sentence pairs spanning 44 languages (all paired with English). In other words, none of those 44 languages were truly “zero-shot” for translation, since each had some parallel English examples in the training mix.

Moreover, the incidence of bilingual content strongly correlates ($r \thickapprox 0.94$) with the presence of that language in the corpus. Languages that appear more often in monolingual form also tend to have more bilingual instances and translation pairs. This means high-resource languages like French or Spanish not only have lots of data in PaLM’s training set, but also many incidental translations (millions of translation pairs), whereas a low-resource language (e.g. Telegu, Gujarati) has little of both (mere thousands of translation pairs).

Identification and manual examination of these bilingual and translation-pair snippets was the next step. The majority (around 55\%) of detected bilingual instances were not explicit translations of the same content. Instead, these cases included phenomena like code-switching within a conversation, embedded references or foreign words (book titles, proper names, or quoted phrases in another language inside an English document), or just unrelated multilingual content that happened to co-occur (e.g. on a single web page).

The remaining bilingual instances (roughly 40\% of them) did involve translation relationships. About half of these were \emph{direct translations}: the same sentence or paragraph repeated in English and another language (sometimes formatted as parallel text sections). The other half were \emph{semantically related cross-lingual} texts that were not exact translations but still conveyed overlapping content; for example, an English passage followed by a summary or commentary in French, or vice versa. These findings illustrate that PaLM’s training data contained a spectrum of bilingual signals, from casual mixtures of languages to clear parallel translations.

To understand how PaLM might leverage these, the study looked at how the parallel sentences were presented. Often, bilingual texts were formatted with explicit or implicit prompts indicating language directions. A common pattern was the use of language labels followed by a colon, such as an English sentence prefixed with ‘\textcolor{blue}
{\texttt{English:}}’ and its translation prefixed with ‘\textcolor{red}{\texttt{French:}}’. This colon-delimited language name format is in fact the default style used in MT research, and here it appears naturally in the data.

Briakou and colleagues also observed variations on this theme: some documents used ISO language codes (e.g. ‘\textcolor{blue}
{\texttt{EN:}}’ or ‘\textcolor{red}{\texttt{FR:}}’) instead of full names, while others used language names in the native tongue (e.g., 
‘\textcolor{red}{\texttt{Français}}’ to signal a French sentence). In some cases, the word ‘Translation’ in the target language was used as an indicator; e.g., a French text might begin with ‘\textcolor{red}{\texttt{Traduction:}}’. These are effectively natural prompts that inform a reader (or a model) that the following text is a translation of a preceding sentence.

Exploiting these insights, Briakou et al. constructed prompts that mirror the patterns found in the training data and tested them on PaLM’s translation tasks. The results showed a marked improvement in PaLM’s zero-shot translation performance when using \emph{data-derived prompts}. In particular, for translations out of English, using the newly discovered prompt formats (such as native language labels) boosted the average translation quality by about 14 chrF points compared to a generic prompting approach.\footnote{chrF \citep{popovic_chrf_2015} is a character-overlap $F$-score metric for translation quality; a 14-point gain is substantial.} This indicates that PaLM had indeed learned to respond to those familiar bilingual cues: when the prompt matched patterns it saw during training, it produced significantly better translations. In practical terms, the model’s true translation ability was higher than initially thought. The “emergent” zero-shot performance was underestimated when users didn’t prompt the model in the right way. By prompting in a style consistent with the incidental training examples, PaLM could be coaxed to translate more accurately and consistently; for example, producing output in the correct target language more reliably. This highlights how context and presentation can govern LLM’s behavior: the relevant “knowledge” was latent in PaLM, yet only fully expressed when the trigger phrase matched its ingrained experience. It’s a bit like a person knowing a skill but needing the right cue or reminder to demonstrate it.

There is, of course, no guarantee that the translation pairs mined from the internet are good. In fact, the general quality of the internet’s multilingual content is low \citep{kreutzer_quality_2022} and continues to degrade due to the rapidly increasing contamination from MT and LLM output \citep{thompson_shocking_2024, kocyigit_overestimation_2025}. Briakou et al. performed an \emph{extrinsic} evaluation of their makeshift corpus of 3.3M French-English sentence pairs extracted from PaLM’s translation instances by using them to train a standard NMT model from scratch and tested its performance. The resulting scores (37–38 BLEU) were not far from those (41 BLEU) for the same NMT trained on the full 40M-sentence WMT dataset. This means that PaLM’s training corpus implicitly contained a parallel corpus sufficient to train a decent translation system. The “needles” it swallowed were largely real and useful translation signals, not just noise. This reinforces the notion that PaLM had effectively learned from legitimate translation examples, so its translation ability is built on a foundation of authentic bilingual knowledge gleaned from the web.

To directly assess how much those incidental translation examples contributed to PaLM’s capabilities, the authors performed \emph{ablation} studies with the model’s scaled-down versions (1B and 8B) and trained them on a \emph{filtered} version of the original PaLM’s corpus. In the filtered data, all of the previously identified translation pairs were removed, simulating a world where the model sees no direct parallel sentences during training. They then compared the translation performance of these models to baseline models trained on the unmodified data.

The ablation results were remarkable: removing the translation pairs caused a significant drop in translation quality, especially for the 1B model. For a set of high-resource language pairs, the 1B model’s average zero-shot translation into English suffered about a 7.4 BLEU-point reduction, and even with few-shot examples the performance remained $\thickapprox{5.9}$ BLEU points lower than the baseline model. This sizable degradation confirms that the small model had been heavily relying on those incidental parallel examples to learn how to translate. In contrast, the larger 8B model also showed an impact but a smaller one: roughly a 2–3 BLEU point drop on the same translation tasks without the bilingual data. The translation ability of the 8B model was hurt by the removal, yet it retained more of its capability than the 1B model did. This trend---a larger relative impact on smaller models, and a diminishing but still noticeable impact on a bigger model---suggests that as model capacity grows, it can compensate to some extent for the lack of explicit parallel data. In other words, a really large model like PaLM might infer cross-lingual mappings from other signals (such as named entities, similar context across languages, and other heuristics) and general language understanding, but the presence of parallel sentences gives a smaller model an irreplaceable head-start in learning to translate. Even at 8B, the fact that performance drops when translation pairs are removed underscores that those incidental examples are indeed a \emph{cause} (not just a coincidence) of the model’s translation skill.

At the same time, the 8B model trained on \emph{English-only} data (no foreign text \emph{at all}) was not found to be completely clueless. It could still translate a bit for languages that use the Latin script, achieving BLEU scores in the teens and twenties for translating \emph{into English}. How is that possible? Likely because some traces of other languages slipped through the filtering, or the model picked up names and tokens that overlap with English as useful seed lexicon “anchors” (Section \ref{x-ling-representation-alignment} above). For example, the model might know some Spanish words (like ‘universidad’) from English context sentences that mention them and thus can map a simple Spanish sentence to English by recognizing shared terms. It might also exploit the fact that languages sharing script have similar character patterns. The takeaway is that a larger model can infer translation mappings from very sparse data---an intriguing hint that at sufficient scale, cross-lingual abilities can emerge from even minimal exposure.

Despite the inevitable limitations\footnote{Even Google researchers could not retrain a full-scale LLM from scratch multiple times!} and numerous advancements that have happened in the last two years, Briakou et al.’s findings invite us to revisit the notion of “emergence” in generative AI. PaLM’s (and other LLMs’) translation skills might initially look emergent as if the capacity to translate popped up out of sheer model complexity and general-purpose learning. However, Briakou et al.’s study reveals that the capability was \emph{latent} in the training data all along. PaLM wasn’t explicitly tasked with translation, but the training data contained the right examples and cues, allowing the model to implicitly learn the task. In other words, the model’s knowledge of translation was acquired through the structure of its experience, not through an explicit directive. This is analogous to a child who grows up in a bilingual household. They were never formally taught to translate, but by constantly hearing two languages side by side, they naturally learn to convert between them. Digesting the internet, LLMs effectively encounter a multilingual world where translations are sometimes provided, and they absorb that correspondence.

The notion of incidental bilingualism emphasizes how learning can occur as a byproduct. For AI researchers, this highlights the importance of carefully examining training data. Many capabilities of LLMs might trace back to hidden lessons in the data rather than purely novel reasoning. This raises the question: should we attribute an ability to the \emph{architecture and learning algorithm} of the model, or to the \emph{information embedded in its environment}? In PaLM’s case, the environment (i.e. the training corpus) contained cross-lingual mappings; the model’s architecture allowed it to pick up on those mappings and store them. So the answer is a bit of both: the capacity to detect and use such patterns is an emergent property of a sufficiently powerful sequence predictor, but the specific skill (French $\leftrightarrow$ English translation) is grounded in having seen actual examples.

Another key insight is that even a small fraction of data can exert a disproportionate influence on behavior. Less than 1\% of PaLM’s training instances were explicitly bilingual, yet those bits had an outsized impact on the model’s performance on translation tasks. This asymmetry points to the efficiency of pattern-learning in LLMs: when the model encounters a rare but highly informative pattern (like a sentence and its translation), it can leverage it to handle many other instances at inference time. It also means that omitted data could lead to omitted capabilities. If, hypothetically, PaLM’s (and, by extrapolation, other full-scale LLMs’) crawl of the web had missed all bilingual content, it might have shown far weaker translation abilities, at least until reaching enormous scale where other heuristics kick in. Thus, the structure and composition of training data directly shape the emergent “translation knowledge.”

Finally, this work has practical implications for improving and understanding LLMs. By recognizing that LLMs may have “translation knowledge” locked behind suboptimal prompting, we may be able to unlock better performance simply by changing how we query the models. This suggests a broader principle: when an LLM seems to lack skill in something, it may be a matter of not accessing the skill properly. The model might be like a library full of facts with no index. If we find the right prompt “index,” suddenly the knowledge comes to the surface. For translation, the “index” could be a native prompt format. For other tasks, there might be analogues (perhaps certain tasks were also seen incidentally in training, and the key is to find the trigger that activates that capability).

In conclusion, “Searching for Needles in a Haystack” paints a narrative of discovery within LLMs’ training data: a substantial translation curriculum might be hidden in plain sight. It shows how an LLM can acquire what looks like new competence from unlabeled data, simply by virtue of the data’s internal structure and diversity. This blurs the line between supervised and unsupervised learning: LLMs are not provided with labeled translation pairs in a traditional sense, but they encounter them in context and may learn from them in a self-supervised way.

But this cannot be the whole story about the origins of LLMs’ translation abilities.

\subsection{Global and Local Learning}

Despite the presence of trace amounts of translation pairs in the non-overlapping token windows---chunks of text consumed by LLMs in single forward passes during pretraining---and their significance in generating LLMs’ translation abilities, most of the multilingual content in the training data comes in the form of monolingual documents found in different parts of the internet. Documents such as news articles or Wikipedia pages in different languages are generally expected to contain semantically identical or closely similar sentences. However, these sentences do not come neatly aligned in pairs, and they are unlikely to appear together within a single sliding context window used during pre-training (typically 4–100K tokens at the time of writing).

We have already noted (Section \ref{how-llms-translate}) that LLMs are capable of learning shared multilingual or language-neutral representations during training on semantically related monolingual documents, thanks to the awesome power of distributional semantics. In that respect, \emph{Global multilingual learning} in LLMs is similar to cross-lingual representation alignment in mBERT \citep{devlin_bert_2019}, mBART \citep{liu-etal-2020-multilingual-denoising}, and its variants,\footnote{And, indeed, in earlier, pre-LLM static-embedding settings; e.g. \citealt{mikolov2013exploitingsimilaritieslanguagesmachine} exploited the potential of efficient learning of linear mappings between the monolingual embedding spaces in simple frameworks such as Skip-gram and CBOW. I am grateful to Michael Carl for reminding me of this important historical precedent.} but on steroids. For example, real-world facts (dates, names, places) tend to appear repeatedly across languages, so the model continuously refines how these concepts are expressed in their respective monolingual contexts. If an LLM sees an article in English about climate change and another in German about ‘Klimawandel’ with overlapping terminology and structure, it can infer \emph{Global alignment} between them. In general, when it encounters a text in language A that is very similar to a text in language B it saw earlier, the only way to reduce perplexity (prediction error) is to internally link those two contexts. Over many such occurrences, the model might cluster representations of sentences by meaning rather than by language. Importantly, this kind of Global learning (alignment, clustering) does \emph{not} happen as a result of scanning single translation instances and adjusting the attention and other model weights in response. Rather, it happens in the course of repeated readjustment of the weights throughout multiple training steps.

Of course, the model occasionally stumbles upon “needles in a haystack”---training instances containing exact translation pairs. When this happens, the model may exploit the multilingual embeddings already learned from the previous steps to better align bilingual context present in a current \emph{Local} context window. And then it may, in turn, leverage this \emph{Local alignment} to improve its global cross-lingual representation potential at the next step.

This kind of ongoing \emph{iteration} between Local and Global learning is a natural and very helpful consequence of \emph{batch training}, in which the model updates its weights after every step. In stochastic mini-batch training, each step’s gradient is computed over a shuffled mini-batch that typically mixes examples from different languages and domains. When a batch happens to contain both an occasional local bilingual cue and many globally related monolingual fragments, the single update aggregates these signals, simultaneously strengthening direct mappings and broader cross-lingual structure. Because mini-batches (often hundreds to thousands of sentences) may be reshuffled and resampled across successive steps and epochs, these heterogeneous mixtures can recur with different neighbors, so weights adjusted on one step are immediately put to work in the next one, closing the loop between Local and Global learning. Once updated, the weights are used in the next iteration, continuously improving the model’s translation abilities (among other abilities it may learn in the process). Notably, the \emph{same local processes} are consistently employed across all stages of training.

In summary, then:

\begin{itemize}
    \item \textbf{Local learning}: acquisition from bilingual signals that co-occur within a single training context window (e.g., an English sentence followed shortly by its translation).
    \item \textbf{Global learning}: alignment of semantically related monolingual content distributed across the training corpus, not necessarily co-occurring in a single window.
    \item The two \textbf{interact iteratively} during batch training.
\end{itemize}

The strategy of continuous iteration between “local” and “global” processes is deeply embedded in machine learning, optimization, and probabilistic modeling. This paradigm allows systems to gradually refine local parameters based on global structure or objectives, and vice versa, converging on a stable final state. The original Expectation-Maximization (EM) algorithm is a foundational framework for this interplay \citep{dempster_maximum_1977}. Many other techniques---from Latent Dirichlet Allocation \citep{blei_latent_2003} to statistical MT \citep{brown_mathematics_1993} to SentencePiece \citep{kudo_sentencepiece_2018}---borrow this pattern.

So, on the balance of evidence and common wisdom, LLMs’ translation skills seem to emerge from a combination of these \emph{Local} and \emph{Global} learning processes. Over the full course of training, the model appears to exploit direct local correspondences when available, and to rely on broader global semantic alignment when they are not. This dual-sourced competence can explain why LLM translations sometimes resemble verbatim dictionary lookups and other times read like fluent paraphrases.

Crucially, this duality hypothesis is not just a post hoc narrative. It may yield concrete, testable predictions about LLMs’ \emph{translation behavior}, which may be sensitive to the relative contributions from two learning mechanisms. I outline three types of such empirical predictions below and offer preliminary considerations on how they could be tested in Section \ref{prospects-empirical}.

\subsection{Empirical Implications of Global and Local Learning}

\subsubsection{Style of Translation Outputs} If the model’s properties learned \emph{locally} (memorized bilingual snippets) are driving a particular translation, the output should skew more literal and form-fixed, whereas globally learned cross-lingual representations will be more adaptive. For instance, we expect LLMs to produce standard, idiomatic translations for common phrases that likely occurred in parallel form during training (e.g., proverb-like expressions or technical terms drawn from bilingual glossaries), but to generate more \emph{explanatory} or rephrased translations for novel or rare expressions.

Recent analyses of LLM outputs suggest that their translations tend to be literal for frequent source-target phrase pairs, yet more interpretative for inputs that were unlikely to have direct parallel examples. This aligns with manual analysis: in earlier studies, GPT models often rendered common idioms in a straightforward way but would explicate obscure idioms in the target language when a ready equivalent is not known (a sign of on-the-fly alignment). For instance, Bureau Works’ analysis \citep{bureau_works_we_nodate} found that GPT-3 achieved 90\% accuracy in translating Chinese idioms, outperforming traditional machine translation engines, which often produced literal and awkward translations. Similarly, \citet{tang_creative_2024} demonstrated that GPT-4, when prompted appropriately, could generate high-quality, context-aware translations of East Asian idioms, surpassing commercial translation engines in both faithfulness and creativity.

These findings are further supported by comprehensive evaluations comparing GPT-4 to human translators. \citet{yan_benchmarking_2024} observed that GPT-4 tends toward overly literal translations and exhibits lexical inconsistency in some situations where human translators sometimes over-interpret context and introduce distinctly human “hallucinations” resulting from fatigue. This suggests that GPT-4's translation behavior varies depending on the familiarity of the source phrases, being more literal with common expressions and more interpretative with less familiar ones.

Accordingly, if we construct a test set of sentences half of which contain well-known idioms (present in training data) and half containing novel metaphors or nonce phrases, an LLM should translate the first category more succinctly and idiomatically, but the second with longer, more descriptive phrasing. Such a stylistic divergence would indicate the model is switching between retrieved local translations and generative global reasoning. If no difference is found---for instance, if the model always produces literal outputs or always paraphrases regardless of input novelty---that would challenge the dual-mechanism account.

\subsubsection{Dependence on Model Scale and Data}
Local vs. global learning should result in different scaling behaviors. A smaller LLM (with fewer parameters or less training data) is expected to rely more on memorized translation examples, while a larger model can compensate via broader semantic abstraction. Indeed, the ablation studies by \citet{briakou_searching_2023} (Section \ref{incidental-bl} above) find that removing incidental parallel data causes a dramatic performance drop in a 1B-parameter model but a smaller drop in an 8B model. We predict this trend to extend to even bigger models: as we scale to tens or hundreds of billions of parameters, translation quality will increasingly survive the removal or absence of direct parallel pairs, because larger models should be able to \emph{infer} better mappings through global context. Conversely, very small models might fail to translate at all without seeing explicit examples. In other words, when a model lacks capacity for deep semantic abstraction, it \emph{needs} local bilingual signals; when it has sufficient capacity, it can achieve a form of translation by aligning semantics across languages (analogous to how multilingual BERT finds a shared embedding space). A concrete empirical test here is \emph{scaling ablation}: train or fine-tune a series of models of increasing size on a fixed corpus with and without parallel examples, then measure translation fidelity. The dual-learning theory predicts an interaction: larger models should show relatively smaller gaps between the “with pairs” and “without pairs” conditions (relying on global learning to fill the gap), whereas smaller models will be crippled without explicit pairs. If instead all model sizes suffer proportionally or a large model is equally impaired by removal of parallel data, that would indicate a single-mode (or at least scale-invariant) learning process, contradicting the hypothesis.

\subsubsection{Generalization to Unseen Language Pairs and Domains}
Global semantic learning implies that an LLM can perform translation even between language pairs or domains it never saw aligned in training, by bridging via meaning. By contrast, purely local learning would struggle in truly novel transfers. We therefore predict that zero-shot translation---especially for unusual language combinations or highly specialized topics---is powered mostly by global mechanism. Empirically, this can be tested by looking at how well an LLM translates between languages that had \emph{no overlapping parallel data} in its training set. If global learning is real, the model should still handle such pairs (perhaps by mapping each language to an English-centric semantic space, effectively pivoting internally; see Sections \ref{pivoting-intermadiate} and \ref{do-llms-pivot-on-their-own} above). In fact, previous research on multilingual NMT showed that a single encoder-decoder model can learn to bridge languages through an implicit “interlingua” when direct pairs are absent from its training data \citep{johnson_googles_2017}. For LLMs, we have contemporary support: the latest GPT and Claude models reportedly perform decently on language pairs like Gujarati-Swahili or Bengali-Turkish \citep{enis_llm_2024} even though it is very unlikely they had parallel data for those pairs; they rely on internal representations and possibly English as a latent pivot.

Recent experiments with toy transformer models trained on highly-structured knowledge graph-based multilingual synthetic factual datasets reported in \citet{blum2025rosettastoneunificationforces} further suggest that larger models should better resist ablations of parallel data \emph{if} training mixtures suppress language-as-shortcut signals (e.g., balanced attribute frequencies, tokenization that blurs script cues).

Another area to examine is the models’ capacity for domain and/or stylistic adaptation. An LLM trained on general web text might never see, say, legal contract sentences in both French and English in the same context. Yet, globally, it learns the legal terminology and style in each language. The dual origin view predicts the model can translate legal text reasonably well (by aligning the semantics of legal expressions learned separately in French and English), but if a specific provision was present verbatim in a training bilingual document, the model might output the known translation verbatim. In evaluations, this would appear as high accuracy in common legal phrases (perhaps even outperforming human translators on consistency) but occasional weird literalness or errors on clauses that require more creative lexical adaptation (since global learning does not provide a direct template). Testing on domain-specific parallel sets that were not in the training mix (to our best knowledge) could reveal this pattern. If observed, it would further support the idea that LLMs have both a “memory” for seen translations and a more dynamic ability to translate via understanding.

In sum, treating LLMs’ translation abilities as having dual origins---part “memory-based,” part emergent alignment---provides a plausible explanation for the mixture of fluent generalization and oddly specific translations these models produce. It also aligns with what we know about scaling laws and training data: bigger models behave more “multilingually general,” and small ones cling to their few examples. The three predictions above offer avenues to verify this account. In the next section, I add some details on  how one might design experiments to \emph{operationalize} these predictions and more rigorously test the dual origin hypothesis.

\section{Prospects for Empirical Testing} \label{prospects-empirical}

If LLMs indeed acquire translation ability via two different mechanisms, how can we verify this in more practical terms? Establishing the relative contributions of Local vs. Global learning is not straightforward because both processes are intertwined in a single pre-training outcome. Nevertheless, a combination of carefully crafted experiments, ranging from corpus ablations to model interventions and output analyses, could illuminate the translation behavior of LLMs, which, in turn, could cast more light on the relative significance of Local and Global learning. In this section, I sketch several experimental paradigms for testing the “duality” hypothesis, and discuss practical considerations: data requirements, model scale, and evaluation metrics. The unifying goal is to observe \emph{differential signatures} of local memorization versus global generalization in translation.

\subsection{Selective Ablation of Fine-Tuning Data}\label{selective-ablation}

One direct way to probe the origins of LLMs' translation behavior is to selectively remove or add certain data from an LLM’s training set and examine the impact. As outlined above (Section \ref{incidental-bl}), this “retrospective surgery” approach was pioneered by \citet{briakou_searching_2023} on PaLM’s smaller replicas in a full-scale training regime. Could it be done in a less expensive fine-tuning mode and applied to other, increasingly available open-source models?

For example, one could take an existing pre-trained model, choose a new language pair, and continue training or fine-tuning the model on an additional dataset from which all parallel sentences for a new language pair are ablated, and then do the same with a version of that dataset where those pairs are exclusively retained but all the monolingual portions in the two chosen languages are removed. By comparing these two extremes---“no parallel data” vs. “only parallel data”---one could, in principle, observe differences in resulting translation performance. The dualism hypothesis predicts that a model fine-tuned without any parallel examples would still develop \emph{some} translation ability in a new language pair (via global semantic alignment), but it might translate more slowly or less idiomatically for complex sentences. Conversely, a model fine-tuned only on parallel sentences would certainly learn direct mappings but might struggle on inputs that require world knowledge or cross-sentence context (since the training signal was confined to one-to-one sentence translations). If adding parallel data for, say, Russian-Japanese suddenly causes the model to produce very literal, consistent Japanese translations for Russian inputs (while translation outputs for other relevant language pairs remain unchanged), that would be a strong indication of a local learning effect.

Such controlled fine-tuning experiments could be done on smaller proxy models first, and data could be added incrementally, in both regimes. This might be similar to classic studies in cross-lingual word embeddings where parallel data were incrementally added to chiefly monolingual training to observe alignment effects (e.g. refining mBERT’s semantic space; cf. \citealt{conneau_emerging_2020}). Key data sources for real-world ablation studies include multilingual web crawls like mC4 or Common Crawl derivatives, where one can identify and remove bilingual segments (using language detectors and alignment algorithms). However, smaller corpus resources---from the latest MT benchmarks or recent real-life translation projects (e.g., the Reeve Corpus, \citealt{balashov-etal-2025-translation})---could also be used in the fine-tuning mode. Model sizes should vary: experiments on 1B, 5B, and 10B parameter models can indicate scaling trends.

As for \emph{evaluation}, standard automatic metrics (BLEU, chrF, COMET) would quantify overall translation quality changes, but it’s equally important to perform targeted manual evaluations. For example, after an ablation, test the model on sentences it \emph{previously} could translate to see if those specific mappings vanished (indicating memory erasure), and test on new, compositionally challenging sentences to gauge if general ability remains. Significant divergences between the differently ablated models---especially where one does well and the other fails (a case of “double dissociation”)---would provide evidence for dual learning pathways.

A note of caution is warranted here. The preceding considerations regarding controlled ablation strictly apply to \emph{full-scale} fine-tuning, in which all model parameters are updated under the same regime as in the initial training. This process is computationally expensive, even for relatively small (1–10B parameter) open-source LLMs. In practice, highly efficient Low-Rank Adaptation (LoRA) methods have become standard for fine-tuning across many tasks, including translation.\footnote{See Sections \ref{quality-and-evaluation}, \ref{low-resource-languages}, \ref{in-context-translation}, and references therein.} While the practical value of LoRA and related techniques is indisputable, their use in testing the duality hypothesis outlined above may be problematic. Our objective is to assess the \emph{differential impact} of two types of fine-tuning data—those with and without “bilingual needles”—on the emergence of highly specific translation abilities in fine-tuned LLMs. With LoRA, however, the fine-tuning signal is diverted away from the main model weights (which remain frozen) into a set of additional low-rank matrices. This architectural change precludes informed judgments about the comparative roles of Local and Global learning, as all new learning is now channeled into the adaptation weights. This redirection may produce a “third pathway” for developing specialized translation abilities, one that merits investigation in its own right; but its relevance to the Local-Global learning dilemma remains uncertain. In light of these considerations, full-scale fine-tuning, though more resource-intensive, appears to be the more appropriate approach.

\subsection{Synthetic Tasks and Controlled Probes}\label{synthetic-tasks}

Another strategy is to design diagnostic tasks that disentangle local from global translation capabilities without the need for retraining or even fine-tuning. An intriguing direction is to investigate whether the distinction between local and global mechanisms re-emerges within the framework of \emph{in-context learning}.

One way to approach this question is to provide the model with \emph{synthetic bilingual contexts} at inference time and examine how it exploits these cues. For instance, we could feed the model a prompt that includes a few embedded \emph{parallel examples} for a new language pair (e.g. a mini-glossary or several translated sentences) and then ask it to translate a novel sentence. This is akin to few-shot prompting. If local learning is a distinct operative mechanism, the model should improve or change style given those explicit examples, essentially performing cross-lingual “translation memory” look-up. On the other hand, if it relies mostly on global knowledge, a few extra examples might not change its performance much. By varying the content of these examples (literal translations vs. paraphrastic ones) and observing the output, one can infer the model’s default bias. A model heavily skewed to local learning might even mistranslate by over-following the format of a misleading example (e.g., if given a bad literal translation as a cue, it might mimic that literality). In contrast, a globally oriented model would stick closer to conveying meaning and be less perturbed by small prompt changes. Recent work on prompt-based analysis of LLM translation indicates that certain prompt formats act as “soft switches” for translation style \citep{aldosari_assessing_2025}; this can be leveraged as a probe.

Additionally, one can create \emph{controlled translation challenges}: for example, craft a paragraph in language A, then present the model with a \emph{shuffled} version of the same paragraph in language B (so that all sentences from A have a counterpart in B, but not in order, and maybe with distractors). Ask the model to translate language A sentences to English. A model using global alignment might pick up on the fact that the content in B includes the same ideas in another language and use it as context (like a human translator consulting a reference translation), whereas a purely local approach might ignore the jumbled B text or not realize its relevance. By measuring improvements in translation when the “reference” text in another language is present (even unordered), we can see if the model performs a kind of \emph{on-the-fly bitext mining and alignment} (a hallmark of global learning). This is a bit like giving the model an open-book exam and asking afterwards: did they use the book or not? If an English LLM translating Catalan text benefits from having the same content available in Spanish concurrently, this implies it can align meanings across languages on its own (since we did not explicitly tell it the Spanish text is a translation). Such behavior was unthinkable in older NMT systems, but with attention mechanisms and huge training, LLMs might do it implicitly.

From a \emph{model internals} perspective, one can utilize techniques from interpretability research to find traces of local vs. global processing. For example, \emph{activation probing} could be used to see if certain neurons or attention heads activate strongly when translating content that the model has seen before. If we had an LLM with known training data, we could take a sentence that appeared in training with its translation, present the source now, and compare the network activations to a case where we present a semantically similar sentence that never appeared in the “bilingual needles” during training. Do different heads light up? Does the model attend to different parts of its context or use different layers more intensively \citep{zhang_exploring_2025}? Tools like \emph{integrated gradients} or \emph{activation patching}\footnote{Activation patching was discussed in Section \ref{language-agnostic-representation} above; see also \citealt{heimersheim_how_2024}. On a recent use of integrated gradients to analyze human and machine translation outputs, see \citealt{sizov-etal-2024-analysing}.} might let us “swap in” or suppress certain computations to test their effect on translation output. For instance, one could attempt to ablate the model’s high-level semantic representation (by intervening in middle layers) and see if it can still translate via literal mappings at lower layers, or vice versa. If disabling a semantic circuit causes the model to output word-by-word translations instead of fluent sentences, that would align with the duality hypothesis (pointing to a distinct global semantic circuit). Conversely, knocking out a “memorization circuit” might degrade names and idiom translation while leaving general capability intact. These interpretability probes are speculative but increasingly feasible as researchers map circuits for specific abilities.\footnote{For example, \citet{wang_interpretability_2023} earlier identified a circuit responsible for indirect object identification in GPT-2.} A parallel effort could perhaps target a “bilingual lexicon circuit” versus a “semantic alignment circuit.”

\subsection{Human Evaluation and Error Analysis}

While automatic metrics will remain central, human evaluation can be indispensable in diagnosing how a translation was produced. Professional translators or bilingual speakers could examine model outputs for signs of literalness, creativity, and consistency with source. One experimental setup might involve a blind evaluation where annotators see pairs of translations from two versions of a model for the same input: a baseline model and one fine-tuned with selectively ablated data, as described in Section \ref{selective-ablation} above, or after priming the baseline model with parallel examples (Section \ref{synthetic-tasks}). If annotators consistently notice differences---say, one translation is more “word-for-word” and the other more “natural”---that would reveal the dual modes. One could also ask annotators to label errors in LLM translations by type: omission, addition, literal translation, mistranslation of idiom, terminology error, and other error categories. Alternatively, one could tailor the MQM framework to these purposes \citep{lommel-etal-2024-multi} and experiment with different context spans---the number of preceding and/or succeeeding sentences shown to the annotators \citep{castilho_survey_2025}. If the dual learning view is correct, we might find a bimodal error distribution: some errors look like over-literalness (the model sticking too closely to source form, potentially when it recalls a parallel), whereas others look like semantic drift or hallucination (the model trying to paraphrase meaning but going astray---an over-extension of global reasoning). Such patterns have been anecdotally reported by users and in case studies (e.g., models occasionally insert a plausible sentence that wasn’t in the source---a global inference gone wrong, or translate a proverb too directly---a “local memory” without adaptation). A systematic annotation of these phenomena across many outputs and languages would provide empirical grounding. Moreover, human experts could probe the model through interactive evaluations: for instance, start translating a sentence and see if the model can continue (testing if it aligns with a human translator’s partial output), or ask the model to explain its translation choices. If the model cites a source phrase mapping (e.g. “I translated X as Y because I’ve seen ‘X means Y’ before”) versus a reasoning (“X is similar to an English concept Z, so I rendered it as...”), it might indirectly reveal the influence of memorized pair vs. real-time reasoning. Of course, models can fabricate explanations, but with careful prompt engineering (or using decoder trace techniques), one might get useful signals.

\subsection{Feasible Settings and Metrics}

It’s worth considering what resources and scales are required for these investigations. Full retraining of a GPT-scale model for ablation is of course out of question, but fine-tuning experiments on smaller scale models (up to 7B or 13B parameters) are within reach for academic researchers today. Using open-source LLMs like Llama, Mistral, OLMo \citep{olmo20252olmo2furious}, or GPT-OSS \citep{openai_introducing_2025} as bases would allow extraction of their training data sources to identify inherent parallel content, which is important for designing ablation studies.

Relatedly, this approarch could cast additional light on \emph{data contamination}. For instance, one could compile all known parallel sentences in the RedPajama or Pile datasets and determine what portion of FLORES-200 test sets \citep{team_no_2022} might have leaked; this is similar in spirit to the controlled study of contamination by \citet{kocyigit_overestimation_2025}, who examined how LLM evaluation on MT tasks is impacted when test examples overlap with training data. Leveraging such findings, we can intentionally create “contaminated” vs “clean” training splits and see how translation performance differs, thereby quantifying memorization.

For automatic evaluation, string-based metrics (e.g., BLEU, chrF) can flag potential “memorized translations,” whereas neural metrics such as COMET may be more informative because they better capture differences between adequacy and fluency. For measuring \emph{alignment} resulting from Global learning, one might use cross-lingual retrieval tasks: e.g., compute the similarity of the model’s representations (from a certain layer) for a sentence and its translation. A model strong in global alignment should yield closer embeddings for translation pairs than for unrelated pairs. This could be evaluated by retrieval accuracy or clustering purity.\footnote{An approach used, e.g., by \citet{wang_english_2022}. Cf. earlier work on reference-free neural network-based evaluation of monolingual inter-sentential “cohesiveness” showing the effectiveness of even very simple earlier static embeddings such as Work2Vec \citep{elmakias_oblivious_2021}.} If an ablated model (with no parallel fine-tuning data) still clusters translations together in embedding space, that’s evidence of global semantic alignment surviving without local signals. If a control model with added parallel data shows significantly tighter clustering, that quantifies the contribution of local learning. One might also repurpose the BLEU difference by \emph{shard metrics}: divide a test set into segments where we believe the model had or lacked direct training exposure, and measure quality on each. A large gap would confirm the two modes.

While dissecting an LLM’s training brain is complex, a suite of complementary\linebreak experiments---\emph{corpus surgery}, \emph{synthetic prompting}, \emph{interpretability analysis}, and \emph{human judgment}---can collectively either reinforce the dual-origin theory or reveal a more unitary explanation. The coming years are likely to see such multi-faceted evaluation setups become standard, as researchers strive to understand \emph{why} these models perform as they do. The next section turns to the broader implications of the dualism hypothesis, if it holds true (or even if it merely prompts us to think differently about translation).

\section{Reconceptualizing Translation in the Age of Deep Learning} \label{implications-concept-translation}

The possibility that high-performing translation by LLMs arises from dual origins prompts us to rethink what “translation” even means in this context. Historically, translation has never been a monolithic, one-size-fits-all process, neither for humans nor machines; and the rise of deep learning amplifies this pluralism. In this section, I consider broader theoretical implications of a \emph{dual-origin translation competence}, touching on issues of pluralism in translation theory, emergent linguistic abilities, and the opacity of AI systems. I also draw connections to major shifts in translation studies, to the undelying ideas in classical philosophy of translation, and to the framework of distributional semantics that underlies modern NLP.

\subsection{Pluralism and Duality in Translation}

If LLMs indeed translate via two pathways---local recall vs. global inference, which are differentially informed by the corresponding learning processes---that underscores a broader lesson: there may be no singular \emph{Translation} with a capital ‘T’. This resonates with the ethos of contemporary translation studies which emphasizes pluralism, the idea that what counts as translation can vary by its context, purpose, and agent \citep{snell-hornby_turns_2006, munday_introducing_2022}. Human translators adopt different strategies on the fly: sometimes translating word-for-word (e.g. for legal texts or patent materials where fidelity to terminology is key), other times paraphrasing or explicating (e.g. when localizing a joke, an idiom, or a culturally-specific content for a new audience). The dualism proposal for LLMs suggests a similar internal diversity. Rather than a single unified algorithm, the model might be juggling something akin to the age-old dichotomy of “literal vs. free translation,” but doing so at a sub-symbolic level through patterns in data. From a theoretical perspective, this aligns with pluralist views that there is no one correct way to translate; success can be achieved through multiple different approaches, even within one mind (or one model).

Interestingly, machine translation history itself has swung between extremes: early rule-based systems (and bilingual dictionaries) versus later statistical approaches with no rules, to neural systems that seek deeper semantic transfer. LLMs may be blending these paradigms: one part of the model’s learning process and the resulting behavior at inference time mimics a \emph{bilingual lexicon} lookup (as rule-based systems did), while another part may leverage \emph{semantic recombination}. Embracing this duality could encourage translation scholars and practitioners to see advanced MT not as a singular “mind” but as an ensemble of different competences working in concert. It also invites a comparison to human bilinguals: does a professional translator ever rely on direct recall (memory of a known translation) versus on-the-spot rephrasing? Anecdotally, yes: experienced translators often remember how certain phrases were translated in the past, or consult translation memory tools (Section \ref{brief-history-trans-tech}), which is essentially externalizing the local memory approach, and mix that with fresh composition. Thus, the LLM’s dual approach might be less alien to human practice than it seems, reinforcing a \emph{pluralistic epistemology} of translation.

This pluralism also resonates with classic philosophical treatments of translation. On Quine’s view, multiple “translation manuals” can fit all the behavioral (and textual) evidence equally well; indeterminacy of translation is not failure but a structural feature of the process \citep{quine_word_1960}. Davidson likewise frames translation as \emph{radical interpretation}: recovering meaning by imposing rational coherence (his Principle of Charity) on a speaker’s utterances in context \citep{davidson_radical_1973}. Read through this lens, the local/global duality in LLMs is a computational echo of those frameworks: “local recall” resembles a manual-like mapping, while “global alignment” resembles the Davidsonian drive toward holistic coherence across a broad web of usage. Both modes can yield adequate translations, yet neither is uniquely mandatory. One can think of it as a modern, sub-symbolic instance of pluralism anticipated in classical philosophy of translation.\footnote{See \citet{rawling_routledge_2019} for a recent collection of articles exploring important questions that arise at the interface of classical translation studies and philosophy of language.}

\subsection{Emergent Translation Competence?}

One striking aspect of LLM translation is its \emph{emergent} nature. The model’s ability to translate was not pre-programmed; it arose as a side effect of scale and data. This raises a question: to what extent is translation an emergent property of any sufficiently sophisticated information-processing system? If cross-lingual alignment can be achieved by exposure to multilingual data, then perhaps \emph{translation competence} in general exists implicitly in the fabric of linguistic meaning. Philosophically, this touches on the concept of a universal meaning space, an idea with a long lineage (from Leinbiz’s \emph{characteristica universalis} to the interlingua in MT research; see e.g. \citealt{hutchins_machine_1986}, Ch. 2). Modern distributional semantics gives this idea computational form: LLMs encode knowledge in high-dimensional vectors, and when trained on many languages, those vectors naturally form clusters by meaning rather than by language (at least for high-level abstractions). Empirical studies back this up: as mentioned earlier, mBERT, for example, was found to have aligned semantic representations across languages without any parallel data, a phenomenon noted as an “emerging cross-lingual structure” \citep{conneau_emerging_2020}. Similarly, recent analyses show that certain LLM families have embeddings where tokens from different languages with the same meaning end up nearby, whereas other families’ embeddings cluster by language \citep{brinkmann-etal-2025-large, lindsey_biology_2025}. The fact that such structure can materialize in an artificial neural network suggests that translation is, at least partly, an emergent property of recognizing sameness of meaning across linguistic form.

\citet{quine_word_1960} cautions us against reifying this emergent common space as a uniquely correct “interlingua.” If translation manuals are underdetermined by evidence, then a high-dimensional distributional space may encode \emph{many} near-equivalent alignments compatible with the same corpus. LLMs select among them based on their training mix and objective. A Davidsonian moral \citep{davidson_radical_1973} might be somewhat different: successful interpretation is secured by maximizing truth and coherence across a speaker’s total (or model-internal) corpus. In practice, instruction-following and decoding objectives act as “charity-like” pressures, steering outputs toward globally coherent continuations. Thus, along one dimension of this classical philosophical framework (Quine), we should expect principled underdetermination within the model’s latent alignments; along another (Davidson), we could expect convergence in use, because the system is constantly optimized to produce coherent, truth-cunducive continuations under task constraints.

However, emergence has degrees. The dualism hypothesis suggests that what has emerged in LLMs is twofold: a capacity to detect translation equivalents when they are explicitly present (a kind of \emph{emergent memory} of translation pairs), and a capacity to merge knowledge from different languages into one representation (an \emph{emergent} “interlingua”). This dual emergence complicates our conceptualization. It means an LLM does not always deploy a single clean language-neutral representation; sometimes it relies on more language-specific correlations. In theoretical terms, we might say LLMs challenge the binary of literal translation vs. “transcreation” by doing both, depending on context. And they do so \emph{without specific instructions}, unlike human translators who consciously choose strategies. This raises the following question: are we witnessing a rudimentary form of meta-cognition in the model (implicitly “deciding” between translation strategies based on input)? Or is it simply the probabilistic application of patterns? Either way, the outcome is an agent that can flexibly translate in different modes. In human terms, that would be considered a highly \emph{competent} translator. So, should we credit the model with a form of translation competence? And if so, what does that competence consist in? Perhaps we need to take a cue from translation studies and broaden the definition of competence beyond “knows how to replace words in one language with words in another” (“translation as retrieval”) to “has internalized a cross-linguistic web of meaning that can be traversed in multiple ways” (“translation as inference”). This perspective connects with work in mainstream semantics suggesting that meaning is not tied to any single language, an idea that translation studies pioneers like \citet{nida_toward_1964} and linguists such as \citet{jakobson_linguistic_1959} would agree with in principle, though they never imagined it instantiated in billions of weighted connections.

\subsection{Opacity and Interpretability: Translation as a Black Box}

Despite these considerations, one must acknowledge the overwhelming \emph{opacity} of how LLMs actually translate. Even if we posit two mechanisms, they are not cleanly separable modules inside the network; they are interwoven in gazillions of parameters. Even if we agree that LLMs’ translation abilities originate from an iterative interplay between two process, Local and Global, the task of saying “how” the model translates a given sentence remains extremely difficult, a familiar problem in explainable AI. For translators and linguists used to working with rule-based or even phrase-based systems, this opacity may be unsettling. It harkens back to the late 20th-century debates on whether translation equivalence can be formally defined or whether it is a fundamentally intuitive, context-bound judgment. Quine’s indeterminacy of translation again comes to mind. In a way, the computational opacity of LLMs embodies a version of his underdetermination in mechanistic form. Many distinct internal parameterizations can implement the same input-output behavior (functional equivalence up to reparameterization). With LLMs, the translation process is distributed across layers of matrix multiplications. The duality hypothesis gives us a conceptual handle, but it remains a theoretical abstraction until we can open the black box. Some efforts in interpretability, as we have seen (Section \ref{language-agnostic-representation}), are trying to do exactly this---identifying circuits or neurons responsible for aspects of translation. But even if we map some components, the holistic behavior might evade full detailed understanding. Does this matter? In practical terms, maybe not: if the model works, users and developers might not care whether it leverages those aspects of its abilities that were learned via route A rather than B internally. Yet, for trustworthiness and further improvement, it does matter. For instance, if we know a model is relying too much on local memory, we might worry about data bias or inappropriate translations being regurgitated (e.g., if it memorized a biased translation of a term from a specific source). Conversely, if it relies on global semantic alignment, we might be concerned about hallucinations or meaning shifts, especially for critical texts (like medical equipment manuals where a “creative” translation could be dangerous). Thus, understanding the balance between local and global learning is important for governance of MT systems.

It also connects to a deeper point: LLM translation challenges the notion of a translator’s intentionality. Human translators can explain why they chose a phrase (even if post facto, as in the older \emph{think-aloud protocol} studies; see e.g. \citealt{bernardini_think-aloud_2001}). What about an LLM? We might program it to output a rationale, but that rationale is just another generated text, not an introspective report. This gap has methodological weight as it forces us to ask: can we ascribe concepts like “strategy” or “preference” to an algorithm? The dualism hypothesis suggests that we use such terms by analogy (we can say the model “leans on” one source of knowledge or another), but ultimately these are metaphors. The reality might be that the model has one huge strategy: next-token prediction, and everything else (including translation behaviors) are subsumed under that. We must be careful, then, in attributing human-like decision-making to LLMs. Instead, what we are really observing is a kind of emergent effect that may only \emph{resemble} having two strategies. This recognition should remind us that any reconceptualization of translation via LLMs has to grapple with the pervasive inscrutability of deep learning. Our theories (duality included) are attempts to impose understandable structure on something intrinsically complex.

\subsection{Continuities with Translation Studies and Distributional Semantics}

Retrospectively, it may still be enlightening to place these developments in the context of major turns in translation studies and the rise of distributional semantics in NLP. Translation studies over decades have seen the \emph{linguistic turn} (focus on equivalence and structure), the \emph{cultural turn} (focus on context, power, ideology, per \citet{venuti_translation_2013} and other scholars), and more recently a \emph{technological turn} (examining how tools and automation shape translation, cf. \citealt{rothwell_translation_2023, moorkens_automating_2025}). The dual-origin story intersects with all three in surprising ways. The local learning aspect of LLMs is akin to a linguistic approach: it deals with correspondences between words and phrases, almost like a probabilistic dictionary. The global learning aspect echoes the cultural/functional approach: it’s about conveying meaning and intent across languages, not tied to form. Meanwhile, the very existence of LLM translators is itself a result of the technological turn which encourages scholars to revisit longstanding assumptions. For instance, translation quality used to be discussed in terms of \emph{equivalence} or \emph{acceptability}, but now we see LLM outputs that are fluent yet may conceal odd source-target mapping behavior (like untranslated bits that “slip through” because the model’s training data had code-switching). This challenges evaluators to differentiate between surface fluency and genuine fidelity. It also rekindles an old debate: Is there a universal translation method? Early MT researchers in the 1950s (following Weaver’s memorandum) searched for a universal code or interlingua; statistical MT later said “no, it's all about data: the more data, the better”; now LLMs suggest “actually, both approaches co-exist.” This is oddly reminiscent of the compromise proposed by some 20th-century translation theorists who argued for a mixed approach: that translators should sometimes be literal, sometimes free, as needed. It’s as if the pendulum swung and finally settled in the middle within the AI itself.

As for distributional semantics, the success of LLMs is another triumph of the famous principle (“you shall know a word by the company it keeps”). In a multilingual setting, this principle naturally leads to cross-lingual representation: words that appear in similar contexts across languages end up with similar vectors. LLMs take this to an extreme, with billions of parameters adjusting to ensure that texts in different languages that talk about the same facts yield similar internal activations. This may be reminiscent of the dream of a multilingual “semantic web.” What’s different in the deep learning era is that it’s not a human-crafted cross-lingual web, but an emergent, opaque one. It might be imperfect; maybe the model has a partial interlingual representation intertwined with language-specific channels. But even that possibility pushes us to reconceptualize translation: not as a discrete mapping from Language A to Language B, but as a kind of lightning flash that illuminates a shared semantic ground from two sides. In practice, when we use an LLM for translation, we are invoking this lightning flash---the model’s vast interlingual knowledge igniting to connect our input and output. The dual nature of the learning process (local/global) which generated this behavior means the flash might follow one of two paths through the cloud of representations, one more direct, one more circuitous. For translation studies, this offers a new metaphor and perhaps a real model of what translation could mean when not performed by conscious agents: it is \emph{enacted equivalence} within a complex system, an alignment of probabilities that produces a text we recognize as a translation.

In conclusion of this section, the age of deep learning invites us---translators, researchers, scholars---to reconceive translation beyond the traditional human-centric frameworks. We must account for multiple co-existing methods (even within one model), accept emergence and opacity as part of the game, and borrow insights from both the humanities and computational sciences. This pluralistic, interdisciplinary understanding is not just academic; it will shape how we build and trust future translation systems, and how we train human translators alongside them (e.g., teaching when to trust the “memory” of an LLM vs. when to rely on its “reasoning,” analogous to using a bilingual dictionary vs. paraphrasing an idea).

\section{Conclusion} \label{conclusion}

LLMs have astonished us with their translation abilities, performing at levels often comparable to specialized systems and human translators. In this paper, we offered a dual-origin account of how these models achieve such feats, positing that LLM translation prowess stems from two sources in pre-training: (1) \emph{Local learning} from incidental parallel data (“bilingual needles in the haystack”), leading to effective memorization of specific translations and the ability to mimic translation format, and (2) \emph{Global learning} from distributed semantic alignment across languages, an emergent capability allowing the model to infer translations by understanding. We saw that neither alone is likely to explain the full picture; instead, it is their \emph{interplay} that produces the remarkable breadth and quality of LLM translations. This duality is supported by initial evidence (e.g. ablation studies, output analysis) and is consistent with the scaling behavior and training conditions of LLMs. It also offers an explanation for why LLMs can translate between languages with no direct training pairs: the global semantic matrix provides the bridge.

Our exploration leads to several \emph{open problems}. First and foremost is the need for more direct empirical verification of the local/global split: can we find definitive signatures or neural correlates of each? We outlined experimental approaches ranging from data ablations to interpretability probes (Section \ref{prospects-empirical}), but these require significant effort and careful design. As LLM research progresses, a priority will be to carry out such experiments, perhaps starting with smaller models as proxies. Data transparency will be key here: researchers need better access to what LLMs have seen in training to truly link training data patterns with output behavior. A related open question is how instruction tuning and reinforcement learning post-training interact with these translation capabilities. We argued in Section \ref{minimal-role-of--instruction-tuning} that instruction tuning likely plays a minimal role in \emph{creating} translation ability, but it surely affects how the model \emph{deploys} that ability (e.g., being more user-friendly, or preferring one style of translation over another). Understanding this interaction---essentially, how a small dose of supervised data can modulate an emergent ability---could inform better tuning strategies for specific domains and language directions.

Another open problem lies in extending high-quality translation to low-resource languages and domains. If our dual-origin hypothesis is correct, it suggests two routes to improvement: feed the model more parallel data for those low-resource cases (to boost local learning) \emph{and/or} find ways to enhance its semantic alignment (perhaps via continued multilingual training or smarter tokenization that links languages). As we have seen (Sections \ref{recent-uses-of-LLMs} and \ref{how-llms-translate}; see also \citealt{xu_survey_2025}), there is ongoing work in both directions. An interesting research avenue is to see if \emph{synthetic parallel data} generated by an LLM (using its global knowledge) can then be used to iteratively improve the same model, essentially bootstrapping global into local knowledge. This could help address the \emph{data scarcity} issue in a virtuous cycle, but caution is needed to avoid reinforcing errors (a risk if the model’s initial translations are imperfect).

From an interdisciplinary angle, we see opportunities to bridge translation studies and AI research. Translation scholars could contribute nuanced evaluation methodologies to better assess not just whether a translation is “good,” but \emph{how} it is good or bad. Conversely, AI findings about LLM translation might revive old linguistic debates under a new light; for instance, the notion of equivalence: LLMs provide a testbed for theories like \emph{dynamic} or \emph{functional equivalence} vs. \emph{formal equivalence} in a completely different organism (a machine). Collaboration across fields could yield richer frameworks for thinking about MT evaluation, perhaps incorporating functional adequacy, audience design, and other human-centric criteria into the loop. It may also lead to improved \emph{human-AI translation interfaces}. If we know an AI has dual translation modes, a translator working with it might query it differently (“give me a literal translation first, then a rephrased one”) or debug an output by asking which words it was unsure about, effectively treating the AI as a junior translator with a vast memory and decent intuition.

Finally, we should acknowledge the evolving nature of “translation” itself. As models blur the line between translation, paraphrasing, and summarization (since all are just forms of transforming text), future research might generalize these as part of a broader capability of “language transformation.” Perhaps what we call translation in LLMs is just one facet of a more general text-to-text transformation ability that also includes rewriting in the same language, explaining in simpler terms, converting code to pseudocode, and so forth. Viewing LLMs through this lens might unify knowledge and lead to improvements across tasks. However, it also raises ethical and quality considerations: a model adept at paraphrasing might inadvertently paraphrase when we wanted a strict translation, or vice versa. Thus, control becomes a crucial open issue: how do we reliably prompt or condition LLMs to follow the desired mode of translation? Research into prompt engineering and decoder constraints is ongoing, but not foolproof.

In closing, the advent of LLMs has propelled machine translation into a new era---one where translation emerges spontaneously from general intelligence-like training, and where the distinction between learning by example and learning by abstraction becomes intriguingly blurred. We have argued that embracing a dual-origin perspective can clarify some of the mystery: it provides a narrative that fits the data and suggests concrete ways to verify and apply this understanding. Whether this dualism stands the test of time or gets refined into a more continuum view (perhaps “mostly global with a sprinkle of local” or vice versa), it is clear that the traditional explanations for MT need updating. By bringing together insights from computational experiments, linguistic theory, and translation practice, we can better grasp the phenomenon of translation in the wild---not as a solved problem, but as a continuing story of humans and machines learning to bridge languages in tandem.\\

\noindent\textbf{Acknowledgments:} The author thanks Chris Wendler, Ryan Nefdt, Michael Carl, members of
the Philosophy and Cognitive Science of Deep Learning group, the audience at the AMTA 2025 Virtual: Emerging AI Breakthroughs and Challenges in Translation Automation, the students in the graduate
seminar on Language Translation Technologies: Past, Present, and Future taught at the University of Georgia in Fall 2025, and the reviewers for \emph{Information} for the very helpful feedback on earlier drafts.\\

\noindent\textbf{Funding:} This research was funded by the National Science Foundation Grant No. \href{https://www.nsf.gov/awardsearch/showAward?AWD_ID=2336713}{SES-2336713}.\\

\noindent\textbf{Institutional Review Board Statement:} Not applicable.\\

\noindent\textbf{Informed Consent Statement:} Not applicable.\\

\noindent\textbf{Data Availability Statement:} There is no data associated with this research.\\

\noindent\textbf{Conflicts of Interest:} The author declares no conflict of interest.\\

\bibliographystyle{plainnat}
\bibliography{LLM_TRA_Abilities}

\end{document}